\documentclass{article}[12pt]
\usepackage[a4paper, total={6in, 9in}]{geometry}

\usepackage{amsmath,amssymb,dsfont,amsfonts,amsthm,amscd}
\usepackage{tabularx}
\usepackage{graphicx}
\usepackage{times}
\usepackage{xcolor}
\usepackage{float}
\usepackage[pdfencoding=auto]{hyperref}
\usepackage{url}
\usepackage{array}
\usepackage{bm}
\usepackage{authblk}
\usepackage{natbib}
\usepackage{setspace}
\usepackage{booktabs}
\usepackage{stackengine}
\usepackage{ulem}
\usepackage{rotating}
\usepackage[caption=false]{subfig}
\stackMath

\newtheorem{prop}{Proposition}[section]

\title{Generating Collective Counterfactual Explanations in Score-Based Classification via Mathematical Optimization}

\author[1]{Emilio Carrizosa}
\author[1]{Jasone Ram\'{\i}rez-Ayerbe}
\author[2]{Dolores Romero Morales}
\affil[1]{Instituto de Matem\'aticas de la Universidad de Sevilla, Seville, Spain\\}
\affil[ ]{\tt {\{ecarrizosa, mrayerbe\}@us.es}}
\affil[2]{Department of Economics, Copenhagen Business School, Frederiksberg, Denmark}
\affil[ ]{ \tt{drm.eco@cbs.dk}}

\date{}

\providecommand{\keywords}[1]{\textbf{\textit{Keywords---}} #1}

\usepackage{pdfpages}

\begin{document}

\includepdf{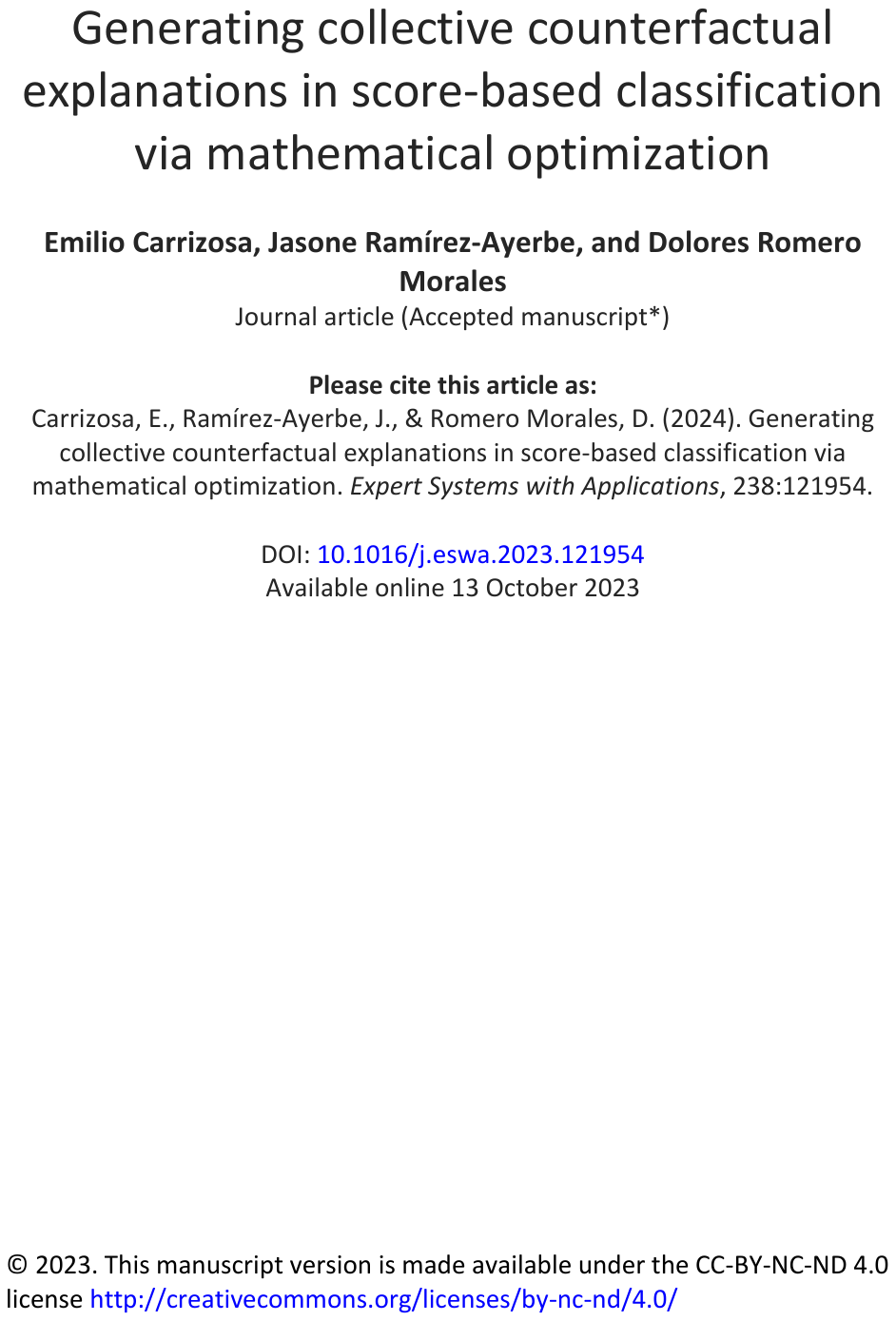}

\newpage 

\maketitle

\begin{abstract}
	
Due to the increasing use of Machine Learning models in high stakes decision making settings, it has become increasingly important to have tools to understand how models arrive at decisions. Assuming an already trained Supervised Classification model,  post-hoc explanations can be obtained via so-called counterfactual analysis: a counterfactual explanation of an instance indicates how this instance should be minimally modified so that the perturbed instance is classified in the desired class by the given Machine Learning classification model.

Most of the Counterfactual Analysis literature focuses on the single-instance single-counterfactual setting, in which the analysis is done for one single instance to provide one single counterfactual explanation. Taking a stakeholder's perspective, in this paper we introduce the so-called collective counterfactual explanations. By means of novel Mathematical Optimization models, we provide a counterfactual explanation for each instance in a group of interest, so that the total cost of the perturbations is minimized under some linking constraints. Making the process of constructing counterfactuals collective instead of individual enables us to detect the features that are critical to the entire dataset to have the individuals classified in the desired class. Our methodology allows for some instances to be treated individually, as in the single-instance single-counterfactual case, performing the collective counterfactual analysis for a fraction of records of the group of interest. This way, outliers are identified and handled appropriately.

Under some assumptions on the classifier and the space in which counterfactuals are sought, finding collective counterfactual explanations is reduced to solving a convex quadratic linearly constrained mixed integer optimization problem, which, for datasets of moderate size, can be solved to optimality using existing solvers.

The performance of our approach is illustrated on real-world datasets, demonstrating its usefulness.
\end{abstract}

\keywords{Collective Counterfactual Explanations; Mathematical Optimization; Explainable Machine Learning; Linear Models; Random Forests}

\section{Introduction}
There is an unprecedented need for transparency and interpretability
in Machine Learning, i.e., explaining how models arrive at decisions \citep{ahaniOR21,jung2020,miller2019explanation, zhang2019should}. This includes selecting the features that impact the most the model as a whole \citep{EnsembleCOVIDEJOR21,bertsimasAS16,HazimehOR20,ZhengOR21}, but also locally to the decision made for  a given
individual \citep{lundberg2017unified,ribeiro2016should}. Given an already trained Supervised Classification model, an effective class of post-hoc explanations consists of the so-called counterfactual explanations \citep{martensMISQ14,wachter2017counterfactual},
identifying the actions to be taken by an instance
(e.g., increase a given amount  the salary, decrease the current debt of an individual  by a given percentage) such that the Machine Learning model at hand would have classified it in the desired class (e.g., the loan request is granted to the individual). 

Methods to build counterfactual explanations for Supervised Classification can be widely divided into model-agnostic or model-specific. The former treat the classification model as a black-box and do not use its inner workings \citep{dandl2020multi, guidotti2019,Poyiadzi_2020}, whereas the latter do. One of the first model-specific approaches was proposed by \cite{wachter2017counterfactual}, where the problem of finding a counterfactual explanation for an instance is formulated as an optimization model, but the numerical solution approach is limited to differentiable classification models \citep{joshi2019towards, ramakrishnan2020synthesizing}. Other model-specific approaches have been proposed for linear classification models \citep{navas2021optimal,ustun2019actionable}, tree-based models \citep{cuichen,FERNANDEZ2020196, parmentier2021optimal} or neural networks \citep{dhurandhar2018explanations,le2020grace}. When differentiability holds, gradient-based algorithms are often proposed to solve the so-obtained optimization problems \citep{joshi2019towards,le2020grace,wachter2017counterfactual}. Otherwise, it is common to find mixed integer programming algorithms \citep{carrizosacounterfactualF23,Fischetti2018,kanamori2020dace,kanamori2021ordered,maragno2022counterfactual,russell2019efficient}. See \cite{karimi2022survey,verma2022counterfactual} for recent surveys and also \cite{browne2020semantics,freiesleben2022intriguing} for works on adversarial learning, with well-known applications in, e.g., image analysis.

The approaches mentioned above, either model-agnostic or model-specific, are all single-instance single-counterfactual  (one counterfactual is sought for just one instance) models. In this paper we depart from such approaches, and, taking a stakeholder perspective, we focus on the generation of collective counterfactual explanations, where a counterfactual is obtained for each instance of a specific group of interest,  which may represent the people of certain age, gender, race, or just the whole set of individuals who were not classified in the desired class by the Machine Learning classifier. There are different reasons to compute the counterfactual explanations for a group of instances instead of computing them separately. First, stakeholders may want to identify a small set of features that need to be modified to alter the prediction of the instances in a group of interest. In the literature, it is often argued that sparser counterfactuals are desired, as having fewer features changed may be more interpretable for an individual, see \cite{carrizosacounterfactualgroup, miller2019explanation, verma2022counterfactual} and references therein. As we are dealing with a group of instances, on top of counting the number of features perturbed  for each instance, with our methodology we can also take into account the total number of features perturbed,  allowing us to identify which features are critical to alter the class prediction given to a group of interest by the classifier. 

Second, there may be some instances that do not behave like the rest of the group, and by perturbing them we would incur very high costs. These instances may be outliers to the group, and it is important to identify them to be treated separately. Whereas outliers' identification is a well studied topic in data analysis, e.g.\ \cite{boukerche2020outlier}, as far as the authors are aware, this issue has not been addressed so far in the literature on counterfactual explanations.

Assuming that a dissimilarity or a distance in the space of instances is given, we propose a mathematical optimization model that minimizes a cost measure of the dissimilarity between the given instances and the collective counterfactual instances. Desirable properties are accommodated by the model as well. In particular, one can control the number of features to be perturbed in order to have all individuals in the group of interest classified in the desired class. This means, in  the loan application example, to identify  the minimum number of changes needed for the group of interest to be granted the loan request, that prior to perturbing the features were rejected. 
Constraints on the counterfactuals are easily accommodated as well \citep{maragno2022counterfactual,mohammadi2021scaling,wachter2017counterfactual}. Our model also allows a certain fraction of records to be treated separately and individually, as in the single-instance single-counterfactual case, and the collective counterfactual analysis is thus performed for just a fraction of records of the group of interest. This way, outliers in the group of interest are properly identified and handled appropriately.

Although our approach is applicable to any classification model based on scores, \cite{carrizosaCOR13}, we focus in what follows on models with some linear structure, such as logistic regression or additive trees like Random Forests or XGBoost. Under mild assumptions on the set defining the feasible perturbations, our optimization model is expressed as a  Mixed Integer Convex Quadratic Model with linear constraints, which can be solved with standard optimization packages.

The remainder of the paper is organized as follows. In Section \ref{sec2}, we formalize the model and introduce the optimization problem that obtains collective counterfactual explanations for  score-based classifiers. We detail this formulation for a specific cost function, accounting for the perturbations as well as the number of features changed, and for two families of classifiers, namely, a logistic regression model and additive tree models. In Section \ref{sec:experiments}, numerical illustrations for real-world datasets can be found. We end the paper with Section \ref{conclusion}, where conclusions and possible lines of future research are provided. The appendix contains a detailed discussion of the formulation for the model for additive trees. The supplementary material contains some of the illustrations from the numerical section.

\section{Problem Statement}
\label{sec2}
\subsection{Collective counterfactuals}
In this section we formalize the problem of finding collective counterfactual explanations. We first discuss the main ingredients of the model, namely, the admissible perturbations, the way perturbations costs are measured, and the baseline classifier.  We then address an easy but relevant case, namely, the separable case, which reduces the problem to solving a series of single-instance single-counterfactual problems. Then, for the non-separable case, a mixed integer programming formulation is given.

Let us consider a binary classification problem, with classes denoted by $+1,-1$, in a set $\mathcal{X} \subset \mathbb{R}^J$. The class $+1$ represents something positive for the individual, such as receiving social benefits or having a loan request granted, while the negative class $-1$ refers to something negative, such as not having access to those benefits or having the loan request denied. Having the data in $\mathbb{R}^J$ does not mean that all features considered are numerical ones. Indeed, if we have categorical features, we can transform them into numerical ones using the standard one-hot encoding.

We assume given a so-called score-based classifier, i.e., one has a function $f: \mathcal{X}  \longrightarrow \mathbb{R}$, so that instance $\bm{x}$ is assigned to the positive class if $f(\bm{x}) \geq \nu$, for a given threshold value $\nu$. For an instance $\bm{x}^0,$ its counterfactual  is an instance $\bm{x}$ in the feasible set $\mathcal{X}^0 \subset \mathcal{X}$, classified in the positive class,
i.e.,  $f(\bm{x}) \geq \nu,$ such that the cost of perturbing $\bm{x}^0$  to $\bm{x}$ is minimal.

Instead of finding the counterfactual for a specific instance $\bm{x}^0$, in this paper we address the problem of building simultaneously collective counterfactual explanations, i.e., explanations for a group of instances.

Let us formalize these ideas and express the search of collective counterfactuals as a mathematical optimization problem.
Let $\bm{\underline{x}}^0= (\bm{x}^0_1,\dots,\bm{x}^0_I)$ be $I$ instances, which form the group of interest,
and let $\mathcal{I}^{*} \subset \{1, \dots,I\}$  be the sought subset of cardinality $I^{*}$ corresponding to the indices of the instances to be perturbed so that the classifier would classify them all in the positive class.
Let $\mathcal{\underline X}^0$ be the feasible region of the set of counterfactuals $\bm{\underline{x}}:=(\bm{x}_1,\dots,\bm{x}_I)$, where $\mathcal{\underline X}^0 \subset \prod_i \mathcal{X}^0_i$ with $\mathcal{X}^0_i$ being the feasible set for the counterfactual of instance $i$, assumed to be a compact set.

The set $\mathcal{\underline X}^0$ includes different constraints that should be satisfied by each single counterfactual \citep{maragno2022counterfactual,mohammadi2021scaling,wachter2017counterfactual}. The set can be a finite collection of datapoints, yielding the so-called endogenous counterfactuals \citep{smyth2022few, wiratunga2021discern}, or defined as a convex combination of known datapoints \citep{brughmans2023nice}. Alternatively, the counterfactual can be synthetically built, yielding in this case the so-called exogenous explanations. The latter is the most popular approach in the literature \citep{guidotti2022counterfactual}. Further constraints may have to be satisfied, for instance, some features such as gender or race cannot be perturbed, for some features such as salary we need to ensure their non-negativity, while for categorical features such as employment status exactly one category should be chosen for each counterfactual instance. In addition to these individual constraints, constraints that may link the counterfactual instances $\bm{\underline x}$ can also appear naturally. For instance, for a categorical feature, we may want that the counterfactuals for all the instances in the group of interest are evenly distributed across all the categories. Doing so, one could avoid unrealistic scenarios, such as requiring all individuals to be in the highest income bracket.

Let $C(\bm{\underline x}^0,\bm{\underline x})$ be the function that measures the cost of perturbing $\bm{\underline x}^0$ to $\bm{\underline x} \in \mathcal{\underline X}^0.$ Although our approach is valid for more general settings, we will consider in what follows a parametrized family of cost functions $C:$
\begin{equation}
	\label{eq:generalC}
	C(\bm{\underline{x}}^0,\bm{\underline{x}})= \sum_{i=1}^I \|\bm{x}^0_i-\bm{x}_i\|_2^2 +  \lambda_{ind}  \sum_{i=1}^I \|\bm{x}_i^0-\bm{x}_i\|_0+\lambda_{glob} \gamma_0(\underline{\bm{x}}^0,\underline{\bm{x}}).
\end{equation}
The first term in \eqref{eq:generalC} is an overall measure of the perturbations (measured through the Euclidean distance) to move from $\bm{\underline{x}}^0$ to $\bm{\underline{x}}.$ The second term measures, weighted by  $\lambda_{ind}\geq 0$, the overall number of perturbations in features needed to move from $\bm{\underline{x}}^0$ to $\bm{\underline{x}}.$ Finally,
$\gamma_0$ is given by
\begin{equation}
	\label{eq:l0def}
	\gamma_0(\underline{\bm{x}}^0,\underline{\bm{x}}) = \left\|\left(\max_{1 \leq i \leq I} |x_{ij}^0-x_{ij}|\right)_{j=1}^J\right\|_0,
\end{equation}
where $x_{ij}$ denotes the value of feature $j$ of instance $\bm{x}_i$. Notice that $\left(\max_{1 \leq i \leq I} |x_{ij}^0-x_{ij}|\right)_{j=1}^J$ is a vector of $J$ components, where each component is the maximum change across all the instances. Hence, the third term measures, weighted by $\lambda_{glob} \geq 0,$ the number of features perturbed ever to move from $\bm{\underline{x}}^0$ to $\bm{\underline{x}}.$

\subsection{Problem formulation}
Given the score-based classifier induced by the score function $f,$ the feasible set $\mathcal{\underline X}^0$ and the cost function $C$ as in \eqref{eq:generalC}, the mathematical optimization formulation of finding the minimal cost perturbations of the features to classify $I^{*}$ instances in the positive class reads as follows:
\begin{align}
	\small
	\min_{\bm{\underline{x}},\mathcal{I}^{*}} \quad  &C(\bm{\underline{x}},\bm{\underline{x}}^0) 	
	\tag{ColCE}
	\label{eq:generalP}\\
	\text{s.t.} \quad &f(\bm{x}_i) \geq \nu \quad \forall i \in \mathcal{I}^{*} \label{eq:cons_score}\\
	&\bm{x}_i=\bm{x}_i^0 \quad \forall i \notin \mathcal{I}^{*} \label{eq:cons_i1}\\
	&|\mathcal{I}^{*}|=I^{*} \label{eq:cons_i2}\\
	&\bm{\underline x}\in \mathcal{\underline X}^0 \label{eq:cons_plaus}.
\end{align}

Let us first address an easy yet very relevant case. When  $\lambda_{glob}=0,$ i.e., when the perturbation cost is measured by the sum of the squared Euclidean distances and the $\ell_0$ norm between each $\bm{x}^0_i$ and its corresponding individual counterfactual $\bm{x}_i,$ then $C$ is separable. Moreover, when  there are no  linking constraints in $\mathcal{\underline{X}}^0$, i.e., when $\mathcal{\underline X}^0= \prod_{i=1}^I \mathcal{X}^0_i$, then Problem \eqref{eq:generalP} is separable as well. This way, the problem can be split into $I$ individual counterfactual problems, linked just by the constraint \eqref{eq:cons_i2}. This
yields the following:
\begin{prop}
	\label{prop:sep}
	Suppose $\lambda_{glob}=0,$ and $\mathcal{\underline X}^0= \prod_{i=1}^I \mathcal{X}^0_i$.
	For $i=1,\dots,I,$ let $\bm{x}^{*}_i$ be an optimal solution of the single-instance single-counterfactual problem
	\begin{align}
		\small
		\min_{\bm{x}_i}\quad  & \|\bm{x}^0_i - \bm{x}_i\|_2^2+\lambda_{ind}\|\bm{x}_i^0-\bm{x}_i\|_0 \label{eq:CEsep} \tag{CEsep}\\
		\text{s.t.} \quad & f(\bm{x}_i) \geq \nu \label{eq:cons_sep}  \\
		&\bm{x}_i\in  \mathcal{X}^0_i. \nonumber
	\end{align}
	Then an optimal solution for Problem \eqref{eq:generalP}  is obtained by sorting the costs $\|\bm{x}_i^0-\bm{x}_i^{*}\|_2^2+\lambda_{ind}\|\bm{x}_i^0-\bm{x}_i^{*}\|_0$ and selecting the $I^{*}$ instances with the lowest value of such cost.
\end{prop}

In what follows, we address the case in which the separability assumptions above do not hold.
Let us express then \eqref{eq:generalP} in a more tractable form, by first rewriting the objective function as a quadratic convex function, and then rewriting the constraints \eqref{eq:cons_score}-\eqref{eq:cons_i2}, so that, when $\mathcal{\underline X}^0$ is polyhedral (eventually with some coordinates in integer numbers), a mixed integer convex quadratic program is obtained.

In order to linearize the $\ell_0$ norm and $\gamma_0$ in \eqref{eq:l0def}, binary variables are introduced. For every feature $j$ and every instance $i$, define the binary variable $\xi_{ij}$ with $\xi_{ij}=1$ if feature $j$ of instance $i$ is perturbed, i.e., $x_{ij}\neq x_{ij}^0$.
The $\ell_0$ norm is linearized by adding to the model following constraints:
\begin{align}
	-M_{ij}\xi_{ij} \leq x^0_{ij}&-x_{ij} \leq M_{ij} \xi_{ij} \quad j=1,\dots,J, i=1,\dots,I \label{eq:l0i}\\
	\xi_{ij} &\in \{0,1\}, \label{eq:l0i2}
\end{align}
\noindent where each $M_{ij}$ is a big constant, which exists thanks to the bounded nature of $\mathcal{\underline X}^0.$ In particular, any 
\begin{equation}
	\label{eq:Mij}
	M_{ij} \geq |x_{ij}^0| + \max_{\bm{z} \in \mathcal{X}^0_i} |z_j|
\end{equation}
is valid.

A similar linearization can be done for the global sparsity $\gamma_0$. In that case, a binary variable $\xi^{*}_j$ is introduced. For each feature $j$, $\xi_j^{*}$ takes the value 1 if $x_{ij}$ takes a value different from $x_{ij}^0$ for some instance $i=1,\dots,I$. The following constraints are added to our formulation:
\begin{align}
	\xi_j^{*} \geq &\xi_{ij}  \quad  i=1,\dots,I, j=1,\dots J\label{eq:l0g1}\\
	\xi_j^{*} &\in \{0,1\} \quad j=1,\dots,J. \label{eq:l0g2}
\end{align}

Moreover, in order to indicate whether an instance is perturbed or not, new binary variables $\bm{y}$ are introduced. For $i=1, \dots,I$, $y_i=1$ if instance $i$ is allowed to be perturbed, and $y_i=0$ otherwise. Then
constraints  \eqref{eq:cons_score} and \eqref{eq:cons_i2} are expressed respectively as
\begin{equation}
	\label{eq:cons_score2}
	y_i(f(\bm{x}_i)-\nu)\geq 0 \quad i=1,\dots,I.
\end{equation}
and
\begin{equation}
	\label{eq:cons_i22}
	\sum_{i=1}^I y_i = I^*
\end{equation}

Notice that when using cost function \eqref{eq:generalC}, as we are minimizing the Euclidean norm, if $y_i=0$, then $\bm{x}_i= \bm{x}_i^0$ automatically, so constraint \eqref{eq:cons_i1} can be omitted.

We end this section by rewriting constraint \eqref{eq:cons_score2} in a more tractable form for the case of a linear classifier, such as the one in the logistic regression.  The formulation for additive tree models (ATM), e.g., Random Forest or XGBoost models, can be found in the Appendix. Additionally, the methodology can be applied to neural networks, specifically to Feed Forward Neural Networks with ReLU activation and other well-known score-based models, see \cite{carrizosacounterfactualgroup}.

In the logistic regression model, the score function is given by $\bm{w}\bm{x}+b,$ and thus constraint \eqref{eq:cons_score2} takes the form:
\begin{equation*}
	(\bm{w}\bm{x}_i+b)y_i  \geq \nu y_i \quad i=1,\dots ,I.
\end{equation*}

We can linearize this new constraint using the usual Fortet linearization technique \citep{fortet}. We introduce a new decision variable $u_i$ to model the product $(\bm{w}\bm{x}_i+b)y_i$. The following constraints need to be added to our model:
\begin{align}
	&u_i \geq \nu y_i \qquad i=1,\dots,I \label{eq:fortet1}\\
	&u_i \leq y_iM_i^{*} \qquad i=1,\dots,I\\
	&u_i \geq - y_iM_i^{*} \qquad i=1,\dots,I\\
	&u_i \leq (\bm{w}\bm{x}_i+b)+(1-y_i)M_i^{*} \qquad i=1,\dots,I\\
	&u_i \geq (\bm{w}\bm{x}_i+b)-(1-y_i)M_i^{*} \qquad i=1,\dots,I\\
	&u_i \in \mathbb{R} \qquad i=1,\dots,I\\
	&y_i\in \{0,1\} \qquad i=1,\dots,I, \label{eq:fortet5}
\end{align}
where each $M_i^{*}$ is a big constant. Using the bounded nature of $\mathcal{\underline X}^{0}$ as before, any
\begin{equation}
	\label{eq:Mi}
	M_i^{*} \geq |b|+\|\bm{w}\|_2\max_{\bm{z}\in \mathcal{X}_i^0} \|\bm{z}\|_2
\end{equation}
is valid.

Hence, the  formulation of Problem \eqref{eq:generalP} for a logistic regression model with cost function \eqref{eq:generalC} is as follows:
\begin{align}
	\label{eq:CELR}
	\tag{ColCELR}
	\begin{split}
		\min_{\underline{\bm{x}}, \bm{y},\bm{\xi},\bm{\xi^{*}},\bm{u}} \quad & \sum_{i=1}^I \|\bm{x}_i^0-\bm{x}\|_2^2+ \lambda_{ind} \sum_{i=1}^I\sum_{j=1}^J \xi_{ij} + \lambda_{glob} \sum_{j=1}^J \xi^{*}_j \\
		\text{s.t.} \quad & \eqref{eq:cons_plaus},\eqref{eq:l0i},\eqref{eq:l0i2}\\ &\eqref{eq:l0g1},\eqref{eq:l0g2},\eqref{eq:cons_i22}-\eqref{eq:fortet5}.
	\end{split}
\end{align}
Assuming $\mathcal{\underline X}^0$ to be a bounded polyhedron with some integer coordinates, Problem \eqref{eq:CELR} is a Mixed Integer Convex Quadratic Model with linear constraints. Therefore, it can be solved using a Mixed Integer Linear Programming (MILP) solver.

\section{Numerical illustration}
\label{sec:experiments}

We will illustrate our methodology using real-world data. We have carried out the experiments in 4 datasets. Table \ref{tab:datasets} reports their name, source of origin, number of instances, number of total features and their type. All these datasets include heterogeneous feature types. 
\begin{table}[H]
	\centering	
	\resizebox{1\columnwidth}{!}{%
		\begin{tabular}{cccccccc}
			Name & Source & Number of instances & Number of features & Continuos & Ordinal & Binary & Nominal\\ \hline
			Boston & Sklearn &  506 & 14 & 13 & 0 & 1 & 0 \\
			COMPAS & ProPublica &  6172 & 5 & 0 & 2 & 2 &1 \\
			Credit & UCI & 29623  &  14 & 4 & 7 & 3 & 0 \\
			Students &  UCI & 395 & 30 & 1 & 12 & 13 & 4 \\
	\end{tabular}}\caption{Datasets used for the numerical experiments}\label{tab:datasets}
\end{table}

In this section, we detail the results for the {\tt Boston housing} dataset \citep{harrisonboston}, which is derived from information collected by the U.S. Census Service concerning housing in the area of Boston. We have accessed it from the scikit-learn library \citep{pedregosa2011scikit}. There are 506 instances and 13 features. The description of the dataset can be found in Table \ref{tab:features}. We have a binary classification problem, where where class $+1$ corresponds to the price of an instance being higher than the median value, and $-1$ otherwise.

\begin{table}[H]
	\centering
	\resizebox{\columnwidth}{!}{%
		\begin{tabular}{c|c|c}
			Variable & Definition & Type \\ \hline
			CRIM   &  per capita crime rate by town & numerical\\
			ZN   &  proportion of residential land zoned for lots over 25,000 sq.ft &numerical\\
			INDUS & proportion of non-retail business acres per town & numerical\\
			CHAS & Charles River dummy variable (1 if tract bounds river; 0 otherwise) & binary\\
			NOX & nitric oxides concentration (parts per 10 million) & numerical\\
			RM & average number of rooms per dwelling & numerical\\
			AGE & proportion of owner-occupied units built prior to 1940 & numerical\\
			DIS & weighted distances to five Boston employment centres & numerical\\
			RAD & index of accessibility to radial highways & numerical\\
			TAX & full-value property-tax rate per \$10,000 & numerical\\
			PTRATIO & pupil-teacher ratio by town & numerical\\
			B & $1000(Bk - 0.63)^2$ where Bk is the proportion of blacks by town & numerical\\
			LSTAT & \% lower status of the population & numerical\\
			MEDV & $+1$ if Median value of owner-occupied homes in \$1000's over the 50th percentile, $-1$ otherwise & target\\
	\end{tabular}}%
	\caption{Description of the features of the {\tt Boston housing} dataset}
	\label{tab:features}
\end{table}

The results for the rest of the datasets are detailed in the supplementary material. 

Two classifiers are constructed, namely, a logistic regression model and a Random Forest with $T=100$ trees and maximum depth of 4. In all the optimization problems solved,  in $ \mathcal{\underline X}^0$ we only impose the binary nature of variable CHAS and lower and upper bounds of each feature to be the minimum and maximum value observed across the 506 observations, respectively.

Solving collective counterfactual explanations for these two classifiers boils down to solving problems \eqref{eq:CELR} and \eqref{eq:CEATM}. In addition, we illustrate the separable case \eqref{eq:CEsep}, in which constraint \eqref{eq:cons_sep} has been rewritten in a more tractable way for the logistic regression model as we have done in \eqref{eq:CELR}, and for the Random Forest as in \eqref{eq:CEATM}. These formulations have been implemented using Pyomo optimization modeling language \citep{hart2017pyomo,hart2011pyomo} in Python 3.7. As solver, we have used Gurobi 9.0 \citep{gurobi}. A value of $\epsilon=1\mathrm{e}{-5}$ has been imposed in formulation \eqref{eq:CEATM} and \eqref{eq:CEsep} for the Random Forest. The values of the big-$M$ constants used are defined in \eqref{eq:Mij}, \eqref{eq:Mi}, \eqref{eq:bigM1} and \eqref{eq:bigM2}. Our experiments have been conducted on a PC, with an Intel R CoreTM i7-1065G7 CPU @ 1.30GHz   1.50 GHz processor and 16 gigabytes RAM. The operating system is 64 bits. All experiments have been solved to optimality. The source code and the data to reproduce all results are available at \url{https://github.com/jasoneramirez/CollectiveCE}.

To visualize the counterfactual explanations, different heatmaps are displayed. In each of them, it is indicated the minimum cost perturbations needed for each instance to change its predicted class.  Each column represents a feature and each row an instance. We present the difference between the original value of the feature and the new one from the counterfactual. A positive value for a feature means that the feature value would have to be higher in order to change the class of the instance, and a negative one indicates that the feature value should be lower.

\begin{figure}[h]
	\subfloat[Logistic regression\label{fig:indLR2}]
	{\includegraphics[width=.48\linewidth]{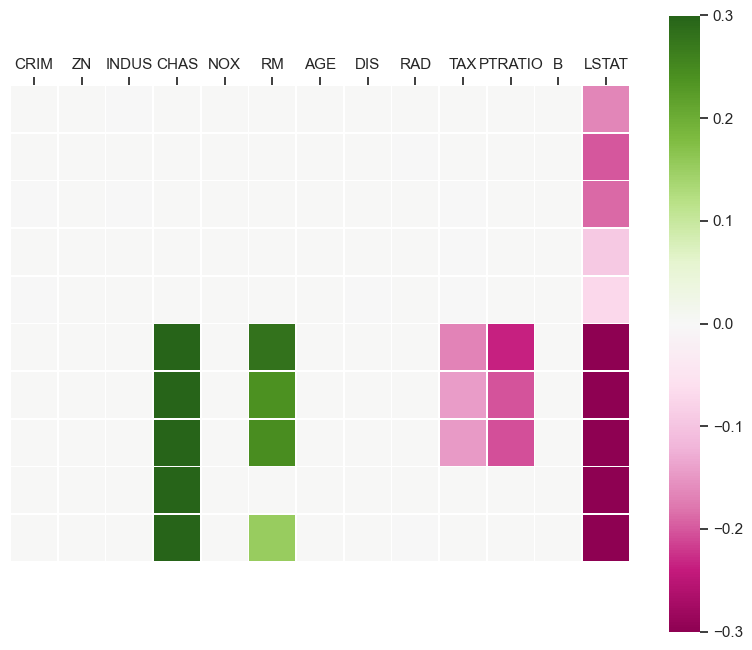}}\hfill
	\subfloat[Random forest\label{fig:indRF2}]
	{\includegraphics[width=.48\linewidth]{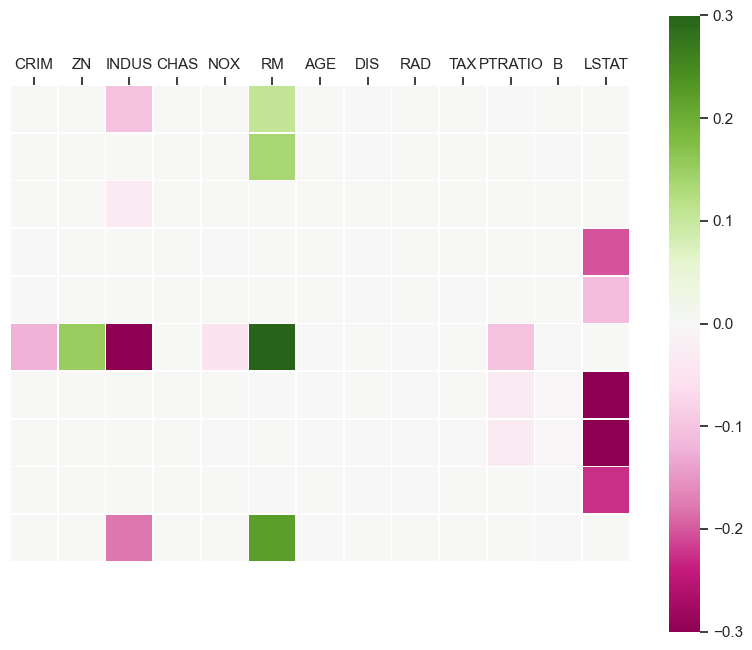}}
	\caption{Counterfactual explanations for instances $\bm{x}_i^0$ in Table \ref{tab:x0values}
		for two classifiers, the logistic regression and the random forest. The explanations have been calculated solving model \eqref{eq:CEsep} with $C_i(\bm{x}_i^0,\bm{x}_i)= \|\bm{x}^0_i-\bm{x}_i\|_2^2 +  \lambda_{ind}  \|\bm{x}_i^0-\bm{x}_i\|_0$ for $\lambda_{ind}=0.02$. The feature perturbations are displayed.}
	\label{fig:CEsep2}
\end{figure}

\begin{figure}[h]
	\subfloat[Logistic regression\label{fig:globLR2}]
	{\includegraphics[width=.45\linewidth]{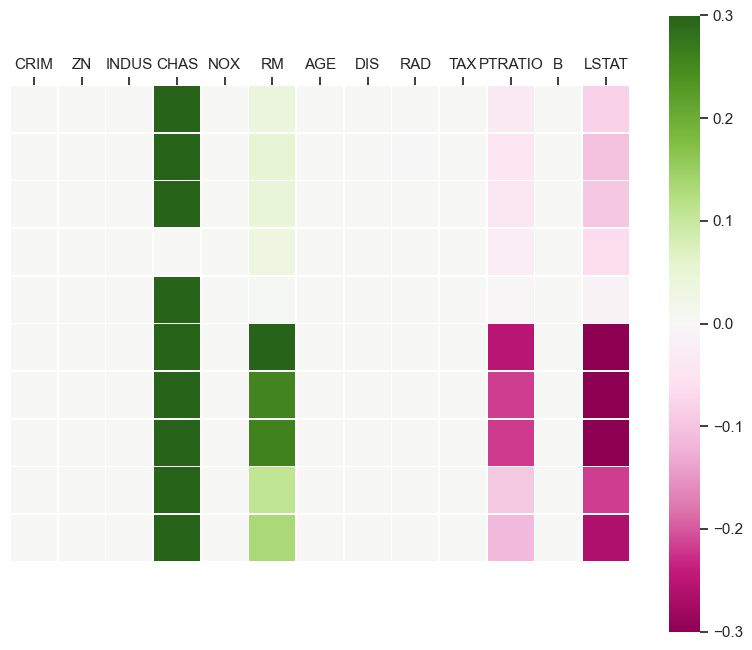}}\hfill
	\subfloat[Random forest\label{fig:globRF}]
	{\includegraphics[width=.45\linewidth]{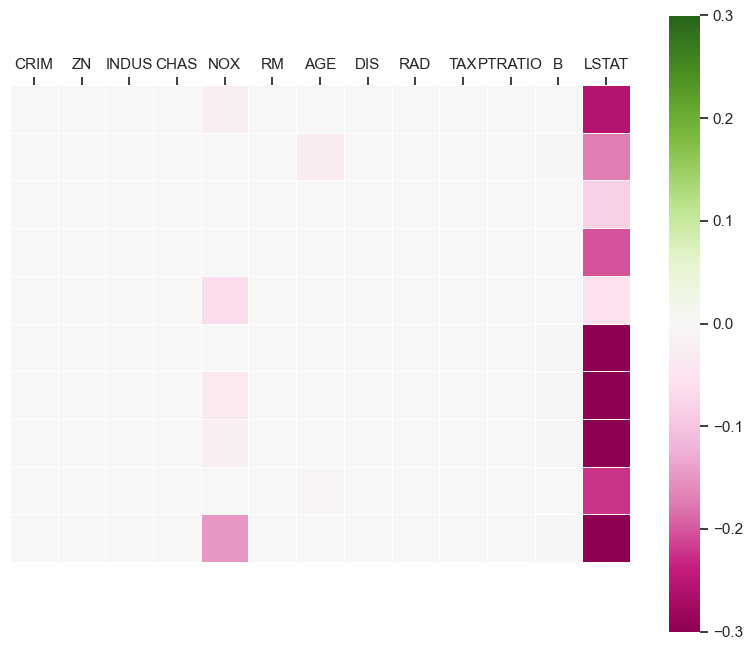}}
	\caption{Counterfactual explanations for instances $\bm{x}_i^0$ in Table \ref{tab:x0values}
		for two classifiers, the logistic regression and the random forest. The explanations have been calculated solving models \eqref{eq:CELR} and \eqref{eq:CEATM} with cost function \eqref{eq:generalC} with $\lambda_{ind}=0$, $\lambda_{glob}=0.2$. The feature perturbations are displayed.   }
	\label{fig:CEgroup2}
\end{figure}

First, counterfactual explanations for 10 individuals that were classified originally in the negative class by both classifiers, are calculated. The original values of the 10 instances can be seen in Table \ref{tab:x0values}. The counterfactuals are calculated with cost function \eqref{eq:generalC} with two different combinations of the values of $\lambda_{ind}$ and $\lambda_{glob}$. Specifically, in the first case, see Figure \ref{fig:CEsep2}, we consider $\lambda_{ind}=0.02$ and $\lambda_{glob}=0$. In such case, as there are no other linking constraints in $\mathcal{\underline X}^0$, the problem is separable, so it is equivalent to solving 10 optimization problems of the form \eqref{eq:CEsep}, see Proposition \ref{prop:sep}. In the second case, see Figure \ref{fig:CEgroup2}, we consider $\lambda_{ind}=0$ and $\lambda_{glob}=0.2$. Now, global sparsity is sought. In all four situations, $I^{*}=I=10$ and we choose $\nu$ such that all the counterfactuals are classfied in the positive class, thus for the logistic regression model $\nu=0$ and for the Random Forest $\nu=0.5$.

\begin{table}[t]
	\centering
	\resizebox{\textwidth}{!}{%
		\begin{tabular}{rrrlrrrrrrrrr}
			\toprule
			CRIM &    ZN &    INDUS & CHAS &      NOX &       RM &      AGE &      DIS &      RAD &      TAX &  PTRATIO &        B &    LSTAT \\
			\midrule
			0.001840 & 0.125 & 0.271628 &  0.0 & 0.286008 & 0.468097 & 0.854789 & 0.496731 & 0.173913 & 0.236641 & 0.276596 & 0.974305 & 0.424117 \\
			0.002399 & 0.000 & 0.236437 &  0.0 & 0.129630 & 0.391071 & 0.608651 & 0.450863 & 0.086957 & 0.087786 & 0.563830 & 1.000000 & 0.399283 \\
			0.001607 & 0.250 & 0.171188 &  0.0 & 0.139918 & 0.417705 & 0.651905 & 0.554320 & 0.304348 & 0.185115 & 0.755319 & 0.995486 & 0.315121 \\
			0.015820 & 0.000 & 0.700880 &  1.0 & 1.000000 & 0.492048 & 0.958805 & 0.056361 & 0.173913 & 0.412214 & 0.223404 & 0.808664 & 0.369481 \\
			0.002870 & 0.000 & 0.346041 &  0.0 & 0.327160 & 0.471738 & 0.901133 & 0.154989 & 0.130435 & 0.223282 & 0.617021 & 0.998487 & 0.275662 \\
			0.124781 & 0.000 & 0.646628 &  0.0 & 0.582305 & 0.257712 & 1.000000 & 0.004056 & 1.000000 & 0.914122 & 0.808511 & 1.000000 & 0.911700 \\
			0.274109 & 0.000 & 0.646628 &  0.0 & 0.648148 & 0.209044 & 1.000000 & 0.030700 & 1.000000 & 0.914122 & 0.808511 & 1.000000 & 0.732616 \\
			0.430994 & 0.000 & 0.646628 &  0.0 & 0.633745 & 0.362522 & 1.000000 & 0.032736 & 1.000000 & 0.914122 & 0.808511 & 1.000000 & 0.796358 \\
			0.137585 & 0.000 & 0.646628 &  0.0 & 0.409465 & 0.436099 & 0.584964 & 0.078931 & 1.000000 & 0.914122 & 0.808511 & 0.061350 & 0.385210 \\
			0.003184 & 0.000 & 0.338343 &  0.0 & 0.411523 & 0.350450 & 0.720906 & 0.151770 & 0.217391 & 0.389313 & 0.702128 & 1.000000 & 0.535596 \\
			\bottomrule
		\end{tabular}%
	}
	\caption{Feature values of $\bm{\underline {x}}^0=(\bm{x}^0_1,\dots,\bm{x}^0_{10})$ used for the models in Figures \ref{fig:CEsep2}--\ref{fig:CEgroup2}.}
	\label{tab:x0values}
\end{table}

It is worth noting how much sense the counterfactuals shown in the different heatmaps make. As one should expect, to increase the value of a house, i.e., change from the negative to the positive class, the counterfactual explanation shows us that the number of rooms (RM) must increase and the \% lower status of the population (LSTAT) has to decrease.

Also, comparing the changes in the separable case (Figure \ref{fig:CEsep2}) with the non-separable case (Figure \ref{fig:CEgroup2}) one must change 5 and 7 features globally in the separable case in the logistic regression case and the Random Forest case respectively, whereas it is reduced to 4 and 3 features in the non-separable case.

Looking at the results, it can be seen that the features LSTAT, RM, CHAS and PTRATIO are the features that have the greatest impact on changing the instances from the negative to the positive class for this group of instances and the logistic regression model. For the Random Forest, the critical features are NOX, AGE and LSTAT.

Secondly, we will display for all the instances in the {\tt Boston housing} dataset that were classified in the negative class by the logistic regression model, i.e.\ that comply with
\begin{equation}
	\label{eq:x0neg}
	\bm{w}\bm{x}_i^0+b<0, \quad \forall i=1,\dots,I,
\end{equation}
the key features that globally would need to change in order to change its class. We consider $I^{*}=I$ and $\nu=0$. To obtain the Pareto front, we will consider the global sparsity term as a hard constraint, instead of in the objective function. Thus, instead of solving \eqref{eq:CELR}, we will consider the following optimization problem:
\begin{align}
	\label{eq:CELRhard}
	\tag{ColCELRhard}
	\begin{split}
		\min_{\underline{\bm{x}}, \bm{y},\bm{\xi},\bm{\xi^{*}},\bm{u}} \quad & \sum_{i=1}^I \|\bm{x}_i^0-\bm{x}\|_2^2  \\
		\text{s.t.} \quad &\sum_{j=1}^J \xi^{*}_j \leq F_{\text{max}}\\
		&
		\eqref{eq:cons_plaus},\eqref{eq:l0i},\eqref{eq:l0i2}\\ &\eqref{eq:l0g1},\eqref{eq:l0g2},\eqref{eq:cons_i22}-\eqref{eq:fortet5}.
	\end{split}
\end{align}

\begin{figure}[h]
	\centering	
	\includegraphics[scale=.5]{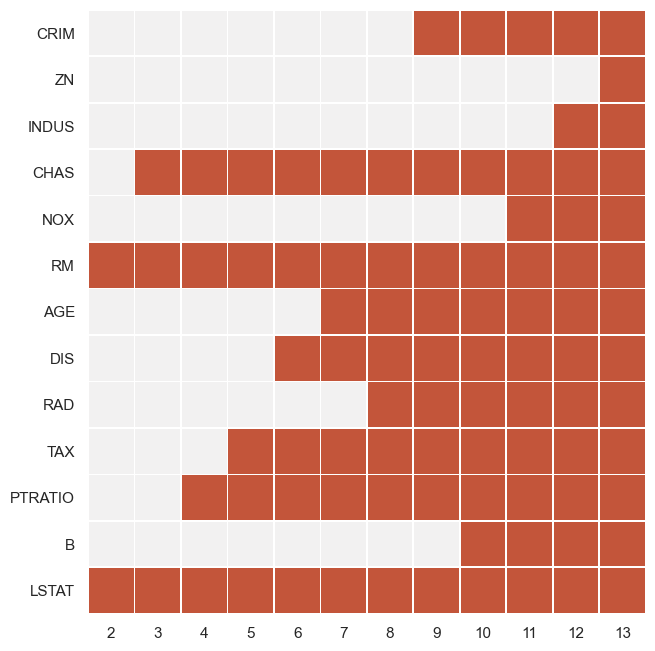}
	\caption{Features that need to be perturbed (in red) for all the instances in the {\tt Boston housing} dataset that comply with \eqref{eq:x0neg} when solving Problem \eqref{eq:CELRhard} for all the values of $F_{\text{max}}$ to describe the Pareto front. The classifier considered is a logistic regression model. $F_{\text{max}}=1$ yields an infeasible problem.}
	\label{fig:featuresLR}
\end{figure}

To describe the Pareto front, we solve this problem for all the possible values of the global sparsity that one can reach, i.e., when $F_{\text{max}}$ takes values 2 to 13. Note that for $F_{\text{max}}=1$, the optimization problem is infeasible. The features that are changed are displayed in Figure \ref{fig:featuresLR}. In this case, we can see a smooth transition in the increase of the perturbed features.

\begin{figure}[h]
	\subfloat[$\lambda_{glob}=0.1$\label{fig:outlier01}]
	{\includegraphics[width=.35\linewidth]{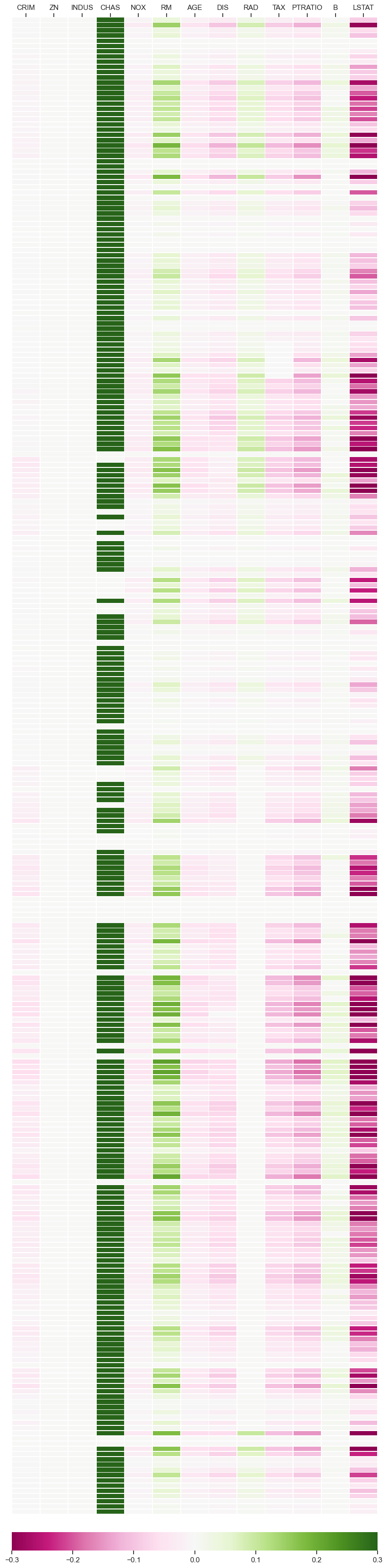}}\hfill
	\subfloat[$\lambda_{glob}=10$\label{figoutlier10}]
	{\includegraphics[width=.35\linewidth]{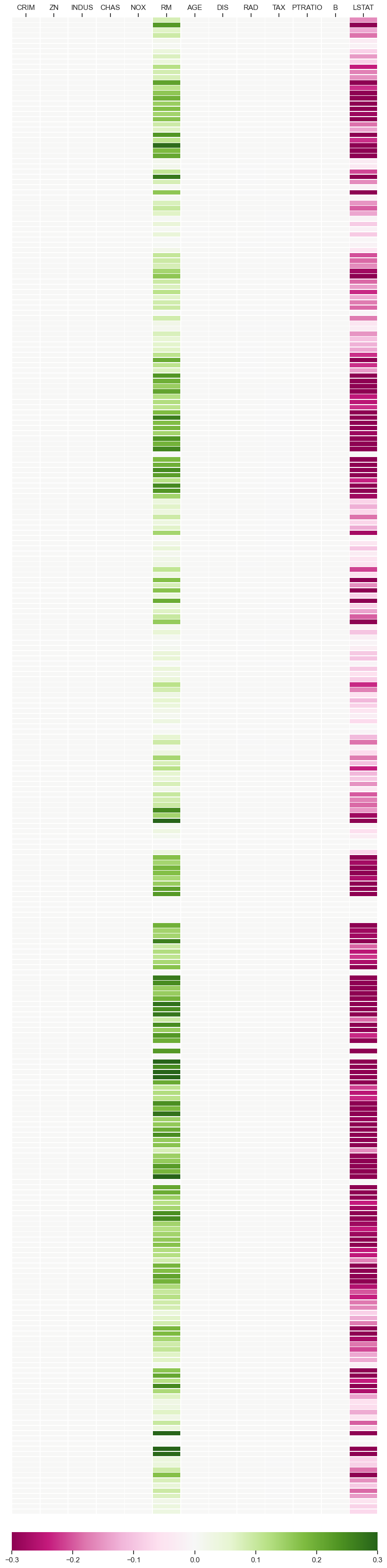}}
	\caption{Counterfactual explanations for all the instances in the {\tt Boston housing} dataset that comply with \eqref{eq:x0neg}. The classifier considered is a logistic regression model. The explanations have been calculated solving model \eqref{eq:CELR} with $\lambda_{ind}=0$, $\lambda_{glob}=0.1$ (left), $10$ (right), and $I^{*}=\lceil 0.95I \rceil$. The feature perturbations are displayed. }
	\label{fig:outliers}
\end{figure}

Thirdly, we look at $\mathcal{I}^*$, which can detect outliers from a group. Considering the same group as before, i.e., the instances that meet condition \eqref{eq:x0neg}, we calculate the counterfactual explanations for 95\% of them, leaving 5\%. We do this for cost function \eqref{eq:generalC} with $\lambda_{ind} = 0 $ and two values of $\lambda_{glob}=0.1,10$. The counterfactuals can be seen in Figure \ref{fig:outliers}.

In this case, for both values of $\lambda_{glob}=0.1,10$ considered the same outliers are outlined. The index of the outliers, i.e., the instances that are not perturbed and therefore do not have a counterfactual associated, are 141, 373, 374, 384, 385, 386, 387, 388, 398, 412, 414, 438, 489, 490.

\section{Conclusions}
\label{conclusion}

In this paper, a unified approach is proposed to build collective counterfactual explanations in classification problems by means of mathematical optimization: minimal cost perturbations on the instances in a group of interest are sought so that, after being perturbed, such instances are classified by the classifier in the desired class. Perturbations costs  are modeled  taking into account both the magnitude of the change and also the number of predictor variables that need to be changed, either individually or globally. The model is expressed as a mathematical optimization problem, detailed for the case of a linear classifier (as e.g.\ in logistic regression or linear SVMs) and for additive trees (as e.g.\  in random forests or XGBoost). We have illustrated our methodology on several real-world datasets.

There are several interesting lines of future research. First, in our numerical illustrations we have certified that all optimization problems have been solved to optimality. However, for problems with a larger number of individuals or a more complex classifier (e.g. a random forest with a large number of trees, or a neural network), it is worthwhile and necessary to investigate more efficient formulations and heuristic algorithms. Second, an extension to our deterministic approach is required when the features measured for instance $\bm{x}^0$ are affected by uncertainty, and thus we have not got any more precise values $\bm{x}^0$ but some sort of uncertainty sets \citep{bertsimasSIAMReview11}. Third, one could have available, instead of a single classification model, a full collection of classifiers, and seek then for a counterfactual explanation leading to the desired change in class assignment in all (or in at least a fraction $\alpha$) of the classifiers, yielding thus a robust counterfactual explanation. Additionally, we are interested in studying the problem for other classifiers, like support vector machine models with non-linear kernels \citep{carrizosaCOR13} or optimal randomized classification trees \citep{blanqueroSORCT20}.  Finally, we have assumed that we can perturb independently the different features not affecting the remaining ones, but, if available, the set $\mathcal{\underline{X}}^0$ could be enriched with causal constraints \citep{pearl2009causality}, so that the perturbation of a feature is restricted by the those of features correlated with it. Accommodating causality in our models is a challenge which deserves further analysis.

\section*{Acknowledgements}

This research has been funded in part by research projects EC H2020 MSCA RISE NeEDS (Grant agreement ID: 822214), FQM-329, P18-FR-2369 and US-1381178 (Junta de Andaluc\'{\i}a, Spain), and PID2019-110886RB-I00 and PID2022-137818OB-I00 (Ministerio de Ciencia, Innovaci\'on y Universidades, Spain). This support is gratefully acknowledged.

\newpage

\bibliographystyle{plainnat} 
\bibliography{bibliografia.bib}

\newpage

\appendix
\section*{Appendix}

\renewcommand{\thesubsection}{\Alph{subsection}}

\subsection{Formulation for ATM}

Consider the case where the classifier is an ATM with $T$ trees. For an instance $\bm{x}_i$, tree $t\in \{1,\dots,T\}$ predicts, with weight $w^t \geq 0$, the positive or negative class. Then, $\bm{x}_i$ is allocated to the class for which the overall weight of the trees classifying $\bm{x}_i$ in such class is maximal. An ATM can be viewed as a classifier based on scores, where the score $f$ is the sum of the weights associated with trees that classify $\bm{x}_i$ in the positive class. More precisely, if $\mathcal{T}_+(\bm{x}_i)$ denotes the subset of trees that classify $\bm{x}_i$ in the positive class, then the score function is defined as:

\setcounter{equation}{23}
\renewcommand\theequation{\arabic{equation}}

\begin{equation}
	\label{eq:scoreATM}
	f(\bm{x}_i)=\sum_{t\in \{1,\dots,T\} : t\in \mathcal{T}_+(\bm{x}_i)} w^t.
\end{equation}

The following notation and decision variables will be used:\\[0.15cm]

\textit{Data}\\

\begin{tabular}{p{0.1\textwidth}p{0.8\textwidth}}
	$\mathcal{L}^t_k$   &  subset of leaves in tree $t$ whose output is class $k=1, \dots, K$, $t=1,\dots,T$\\
	$\mathcal{L}^t$ & set of leaves in tree $t$. Hence, $\mathcal{L}^t=\cup_k \mathcal{L}^t_k$, $t=1,\dots,T$\\
	$\text{Left}(l,t)$  &  set of ancestor nodes of leaf $l$ in tree $t$  whose left branch takes part in the path that ends in $l$, $l\in \mathcal{L}^t$, $t=1,\dots, T$\\
	$\text{Right}(l,t)$ & set of ancestor nodes of leaf $l$ in tree $t$  whose right branch takes part in the path that ends in $l$, $l\in \mathcal{L}^t$, $t=1,\dots,T$\\
	$v(s)$& feature used in split $s$, $s \in \text{Left}(l,t)\cup \text{Right}(l,t)$\\
	$c_s$ & threshold value used for split $s$, $s\in \text{Left}(l,t)\cup \text{Right}(l,t)$\\
	$w^t$ & weight of each tree, $t$, $t=1,\dots, T$\\[0.15cm]
\end{tabular}

\textit{Decision Variables} \\

\begin{tabular}{p{0.1\textwidth}p{0.8\textwidth}}
	$\bm{\underline x}$ & counterfactual, $\bm{\underline x} \in \mathcal{\underline X}^0$\\
	$z_{i,l}^t$ & binary decision variable that indicates whether the counterfactual instance $i$ ends in leaf $l \in \mathcal{L}^t$ ($z_{i,l}^t=1$) or not ($z_{i,l}^t=0$), $t=1,\dots, T$\\[0.15cm]
\end{tabular}

Since the structure of the ATM is given, i.e., the topology of the trees, the splits, and the thresholds used in each of the trees, the only decision variables are those defined to indicate in which leaf the counterfactual $\bm{x}_i$ ends. Once we know in which path the solution is, the branching condition is determined as well. For each split $s \in \text{Left}(l,t)$ the condition is true, $x_{iv(s)}\leq c_s$. Similarly, for each split $s \in \text{Right}(l,t)$ the condition is false, then $x_{iv(s)} >c_s$.

The branching conditions are only activated if we end in the corresponding leaf. We introduce these logical conditions using the big $M$-method:

\begin{align}
	x_{iv(s)} - M_1(1-z_{i,l}^t) + \epsilon \leq c_s \quad s\in \text{Left}(l,t) \label{eq:eps2}\\
	x_{iv(s)}+ M_2(1-z_{i,l}^t) - \epsilon \geq c_s \quad s\in \text{Right}(l,t).\label{eq:eps}
\end{align}

Since a strict inequality is not supported by Mixed-Integer Optimization solvers, following \cite{bertsimas2017}, we have introduced a small constant $\epsilon$ in \eqref{eq:eps2} and  \eqref{eq:eps} and changed the inequality to non-strict.  With this definition, we are not allowing our variable $x_{iv(s)}$ to take values around the threshold value $c_s$ at the split $s$.

Please note that the value of $M_1$ and $M_2$ can be tightened for each split. Assuming all features $j$ are bounded and that the lower and upper bounds are known, i.e., $lb_j\leq x_{ij}\leq ub_j$, $j=1,\dots,J$ $i=1\dots,I$, one can take as big $M$ the following values:

\begin{align}
	M_1=&\max\{|lb_{v(s)}|,|ub_{v(s)}|\}-c_s \label{eq:bigM1}\\
	M_2=&\min\{|lb_{v(s)}|,|ub_{v(s)}|\}+c_s+\epsilon \label{eq:bigM2}.
\end{align}

Moreover, for the dummies of the categorical variables, we can assume that $\epsilon=0$.

Using the decision variables $z_{i,l}^t$ associated with the counterfactual $\bm{x}_i$, we can write the score function as a linear function:

\begin{equation*}
	f(\bm{x}_i)=\sum_{t=1}^T \sum_{l\in \mathcal{L}^t_k} w^t z_{i,l}^t.
\end{equation*}

As before, we will consider the parametrized family of cost functions $C$ in \eqref{eq:generalC}. The formulation for Problem \eqref{eq:generalP} for ATM with this cost function yields as follows:

{\small
	\begin{align}
		\min_{\underline{\bm{x}}, \bm{y}, \bm{z}, \bm{\xi}, \bm{\xi^{*}}} \quad & \sum_{i=1}^I \|\bm{x}_i^0-\bm{x}\|_2^2+ \lambda_{ind} \sum_{i=1}^I\sum_{j=1}^J \xi_{ij} + \lambda_{glob} \sum_{j=1}^J \xi^{*}_j\\
		\text{s.t.} \quad & x_{iv(s)} - M_1(1-z_{i,l}^t) +\epsilon \leq c_s \quad  \forall s\in \text{Left}(l,t)\quad  \forall l\in \mathcal{L}^t \quad  t=1,\dots,T \quad  i=1\dots,I \label{eq:rf1g}\\
		&x_{iv(s)}+ M_2(1-z_{i,l}^t)-\epsilon \geq c_s  \quad \forall s\in \text{Right}(l,t) \quad \forall l\in \mathcal{L}^t \quad t=1,\dots,T\quad i=1\dots,I\label{eq:rf2g}\\
		&\sum_{l\in \mathcal{L}^t} z_{i,l}^t=1    \quad t=1,\dots,T \quad  i=1,\dots,I \label{eq:rf6g}\\
		& \sum_{t=1}^T \sum_{l\in \mathcal{L}^t_{+}} w^t z_{i,l}^t y_i \geq  \nu y_i \quad i=1,\dots,I \label{eq:rf9g}\\
		&\sum_{i=1}^I y_i=I^{*} \label{eq:rfi}\\
		&z_{i,l}^t \in \{0,1\} \quad \forall l \in \mathcal{L}^t \quad t=1,\dots,T\quad i=1,\dots,I\label{eq:rf10g}\\
		&y_i \in \{0,1\} \quad  i=1,\dots,I \label{eq:rf11g} \\
		&\underline{\bm{x}}\in \underline{\mathcal{X}}^0 \label{eq:rf13g}\\
		&\eqref{eq:l0i}-\eqref{eq:l0i2}, \eqref{eq:l0g1}-\eqref{eq:l0g2}. \nonumber
	\end{align}
}%

Constraints \eqref{eq:rf1g} and \eqref{eq:rf2g} determine where the counterfactual instance ends and constraint \eqref{eq:rf6g} enforces that only one leaf is active for each tree. Constraint \eqref{eq:rf9g} enforces that the counterfactual instance is classified in the positive class if it belongs to $\mathcal{I^{*}}$. With constraint \eqref{eq:rfi} the number of individuals to be changed is imposed.  Constraint \eqref{eq:rf10g} and \eqref{eq:rf11g} ensure that each $z_{i,l}^t$ and $y_i$ is binary, and constraint $\eqref{eq:rf13g}$ that the counterfactuals $\bm{\underline x}$ are in the feasible set $\mathcal{\underline X}^{0}$. Constraints \eqref{eq:l0i}-\eqref{eq:l0i2}, \eqref{eq:l0g1}-\eqref{eq:l0g2} are as for Problem \eqref{eq:CELR}.

One can also linearize constraint \eqref{eq:rf9g} using again the Fortet linearization technique. Introducing a new binary decision variable $u_{i,l}^t \in \{0,1\}$. Instead of \eqref{eq:rf9g}, the following constraints are added to our model:

\begin{align}
	&\sum_{t=1}^T w^t \sum_{l\in \mathcal{L}^t_{+}} u_{i,l} \geq \nu y_i \qquad  i=1,\dots,I \label{eq:fort1}\\
	&u_{i,l}^t \leq y_i  \qquad i=1,\dots,I, \quad l\in \mathcal{L}^t_k\quad  t=1,\dots,T \label{eq:fort2}\\
	&u_{i,l}^t \leq z_{i,l}^t \qquad i=1,\dots,I  \quad l\in \mathcal{L}_k^t \quad  t=1,\dots,T \label{eq:fort3}\\
	&u_{i,l}^t \geq y_i+z_{i,l}^t -1\qquad  i=1,\dots,I \quad  l\in \mathcal{L}_k^t  \quad t=1,\dots,T  \label{eq:fort4}\\
	&u_{i,l}^t \in \{0,1\}\qquad  i=1,\dots,I, \quad l\in \mathcal{L}_k^t\quad  t=1,\dots,T. \label{eq:fort5}
\end{align}

The final formulation is as follows:

\begin{align}
	\label{eq:CEATM}
	\tag{ColCEATM}
	\begin{split}
		\min_{\underline{\bm{x}}, \bm{y}, \bm{z}, \bm{\xi}, \bm{\xi^{*}}, \bm{u}} \quad & \sum_{i=1}^I \|\bm{x}_i^0-\bm{x}\|_2^2+ \lambda_{ind} \sum_{i=1}^I\sum_{j=1}^J \xi_{ij} + \lambda_{glob} \sum_{j=1}^J \xi^{*}_j\\
		\text{s.t.} \quad &\eqref{eq:l0i}-\eqref{eq:l0i2}, \eqref{eq:l0g1}-\eqref{eq:l0g2}\\ &\eqref{eq:rf1g}-\eqref{eq:rf6g},\eqref{eq:rfi}-\eqref{eq:fort5}.
	\end{split}
\end{align}

We obtain a Mixed Integer Convex Quadratic Model with linear constraints.

\section*{Supplementary material for ``Generating Collective Counterfactual Explanations in Score-Based Classification via Mathematical Optimization"}

\setcounter{table}{2}
\renewcommand\thetable{\arabic{table}}

\setcounter{figure}{4}
\renewcommand\thefigure{\arabic{figure}}

The same experiments described in Section \ref{sec:experiments} of the main paper for the {\tt Boston Housing} dataset  have been carried out in 3 additional datasets. Table \ref{tab:datasets} of the main paper reports their name, source of origin, number of instances, number of total features and their type. The source code and the data to reproduce all results are available at \url{https://github.com/jasoneramirez/CollectiveCE}.

%

In these datasets, there are some nominal categorical features that have been transformed into numerical ones using the standard one-hot encoding, with one dummy variable for each category. Therefore, in the heatmaps each category is represented. However, one must take into account that a change from one category to another, although visualized as two dummy variables changes in the heatmaps (one is activated and changes from 1 to 0, whereas another is deactivated and goes from 0 to 1), in reality only one categorical feature is being perturbed.

\subsection*{COMPAS}

The {\tt{COMPAS}} dataset was collected by ProPublica \citep{angwin2019machine} as part of their analysis on recidivism decisions in the United States. Following \cite{karimi2020model, Mothilal_2020, parmentier2021optimal}, 5 features are considered. The description of the dataset is detailed in Table \ref{tab:features_compas}. 

\begin{table}[H]
	\centering
	\resizebox{1\columnwidth}{!}{%
		\begin{tabular}{c|c|c}
			Variable & Definition & Type \\ \hline
			Race   &  race of defendants: African-American, Asian, Caucasian, Hispanic, Native American, Other & categorical (nominal)\\
			AgeGroup   &  age category of defendants: 1 ($<$25), 2 (25-45), 3 ($>$45)  & categorical (ordinal)\\
			Sex &  sex of defendants: 0 (male) or 1 (female) & binary\\
			PriorsCount &  prior criminal records of defendants & numerical\\
			ChargeDegree &  charge degree of defendants: 0 (Misdemeanor), 1 (Felony) & binary\\
			TwoYearRecid & binary variable indicate whether defendant is rearrested at within two years: 1 (no), -1 (yes) & target\\
	\end{tabular}}%
	\caption{Description of the features used of the {\tt COMPAS} dataset}
	\label{tab:features_compas}
\end{table}

We will conduct the same experiments done for the {\tt{Boston housing}} dataset. First, counterfactual explanations for 10 individuals that were classified originally in the negative class by both classifiers, logistic regression and random forest, are calculated. The original values of the 10 instances can be seen in Table \ref{tab:x0values_compas}.

\begin{table}[t]
	\centering
	\resizebox{\textwidth}{!}{%
		\begin{tabular}{ccccc}
			\toprule
			Race  & AgeGroup & Sex & PriorsCount & ChargeDegree \\
			\midrule
			African-American&    2 &   0 &           7 &            1 \\
			African-American&     2 &   0 &          16 &            1 \\
			African-American&        2 &   0 &          18 &            1 \\
			African-American&       1 &   0 &           3 &            1 \\
			African-American&       1 &   0 &           1 &            1 \\
			African-American&        1 &   0 &           3 &            1 \\
			Caucasian   &     3 &   0 &          28 &            1 \\
			African-American&     3 &   0 &          38 &            1 \\
			Caucasian    &    3 &   1 &          28 &            1 \\
			African-American    &   2 &   0 &          19 &            1 \\
			\bottomrule
		\end{tabular}%
	}
	\caption{Feature values of $\bm{\underline {x}}^0=(\bm{x}^0_1,\dots,\bm{x}^0_{10})$ used for the models in Figures \ref{fig:CEsep_compas}-\ref{fig:CEgroup_compas}.}
	\label{tab:x0values_compas}
\end{table}

We solve models \eqref{eq:CEsep}, \eqref{eq:CELR} and \eqref{eq:CEATM}, considering different combinations of the values of $\lambda_{ind}$ and $\lambda_{glob}$.

\begin{figure}[H]
	\subfloat[Logistic regression\label{fig:indLR_compas}]
	{\includegraphics[width=.48\linewidth]{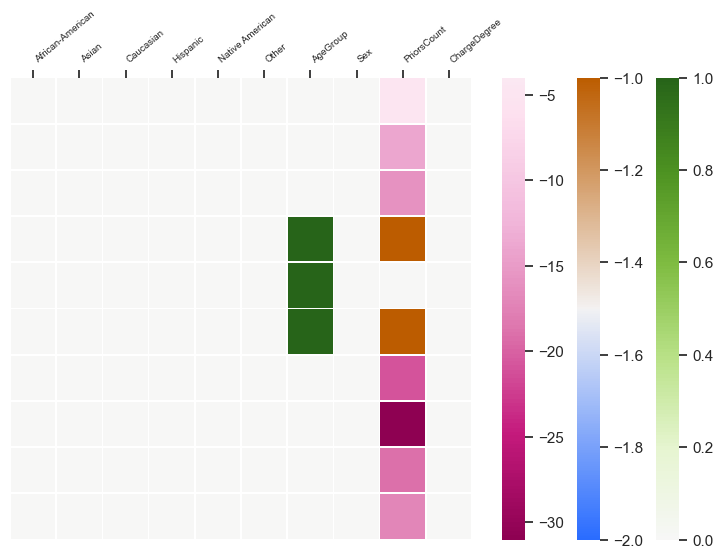}}\hfill
	\subfloat[Random forest\label{fig:indRF_compas}]
	{\includegraphics[width=.48\linewidth]{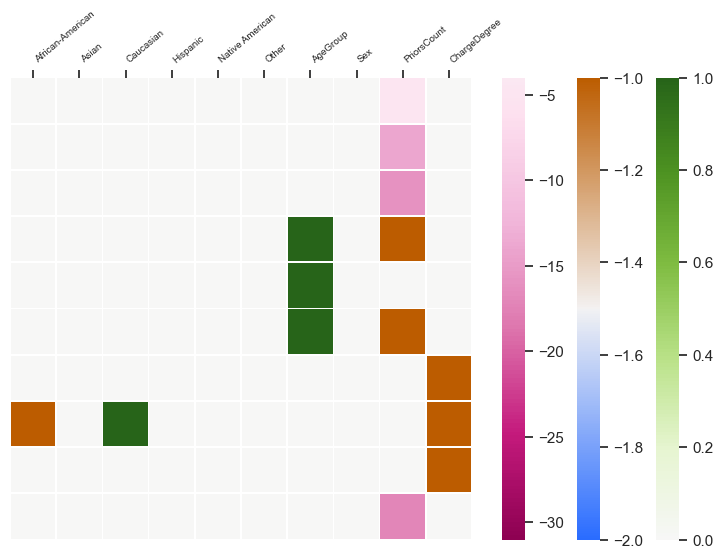}}
	\caption{Counterfactual explanations for instances $\bm{x}_i^0$ in Table \ref{tab:x0values_compas}
		for two classifiers, the logistic regression and the random forest. The explanations have been calculated solving model \eqref{eq:CEsep} with $C_i(\bm{x}_i^0,\bm{x}_i)= \|\bm{x}^0_i-\bm{x}_i\|_2^2 +  \lambda_{ind}  \|\bm{x}_i^0-\bm{x}_i\|_0$ for $\lambda_{ind}=0.02$. The feature perturbations are displayed.}
	\label{fig:CEsep_compas}
\end{figure}

\begin{figure}[H]
	\subfloat[Logistic regression\label{fig:globLR_compas}]
	{\includegraphics[width=.45\linewidth]{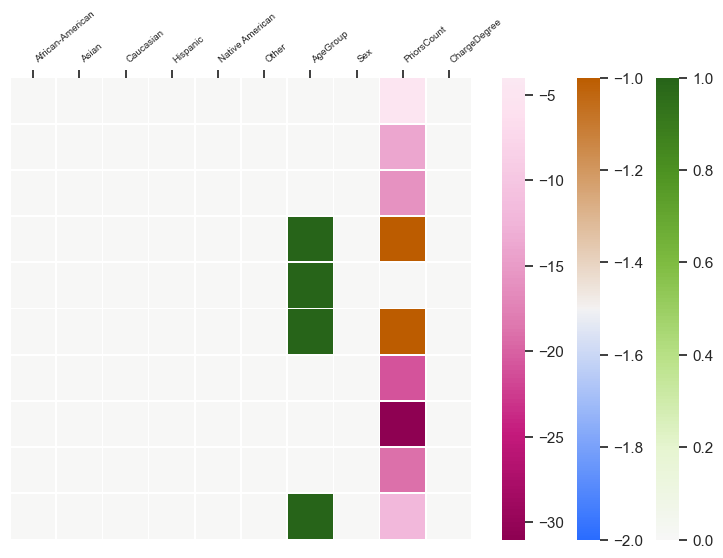}}\hfill
	\subfloat[Random forest\label{fig:globRF_compas}]
	{\includegraphics[width=.45\linewidth]{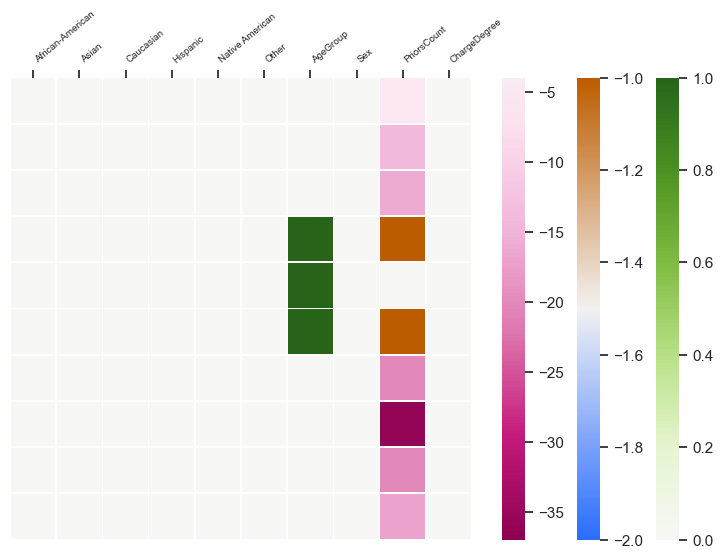}}
	\caption{Counterfactual explanations for instances $\bm{x}_i^0$ in Table \ref{tab:x0values_compas}
		for two classifiers, the logistic regression and the random forest. The explanations have been calculated solving models \eqref{eq:CELR} and \eqref{eq:CEATM} with cost function \eqref{eq:generalC} with $\lambda_{ind}=0$, $\lambda_{glob}=0.2$. The feature perturbations are displayed.   }
	\label{fig:CEgroup_compas}
\end{figure}

For this dataset, the counterfactuals obtained also make sense. As one should expect, to not be labeled as reincident, the number of prior criminal records has to decrease as well as the charge degree. Also, the {\tt COMPAS} dataset has been used to exemplify unfairness in classification. This can be detected, in the separable case, for the Random Forest. Indeed, for one instance, if we were allowed to change  their race from African-American to
Caucasian, decreasing their charge degree would result in a positive classification.

Comparing the results in the separable case (Figure \ref{fig:CEsep_compas}) with the non-separable case (Figure \ref{fig:CEgroup_compas}) we see that 4 features need to be changed globally (i.e., Race, AgeGroup, PriorsCount and ChargeDegree) in the separable case in the Random Forest case, whereas only 2 features need to be changed (AgeGroup and PriorsCount) in the non-separable case. For the logistic regression case, the perturbed features are the same in the separable and non-separable cases. 

Looking at the results, it can be seen that the features AgeGroup and PriorsCount are the features that have the greatest impact on changing the instances from the negative to the positive class for this group of instances and for both models. It is worth noticing that previous work \citep{dressel2018accuracy} about this dataset reached the conclusion that despite having the original dataset a collection of 137 features, the same accuracy could be achieved with a simple linear classifier with only two features. These two features where the age and the priors counts.

\begin{figure}[H]
	\centering	
	\includegraphics[scale=.5]{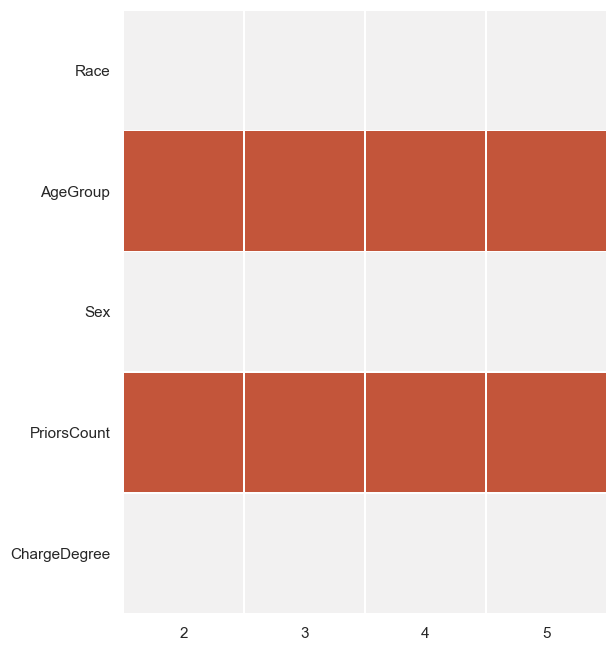}
	\caption{Features that need to be perturbed (in red) for all the instances in the {\tt COMPAS} dataset that comply with \eqref{eq:x0neg} when solving Problem \eqref{eq:CELRhard} for all the values of $F_{\text{max}}$ to describe the Pareto front. The classifier considered is a logistic regression model. $F_{\text{max}}=1$ yields an infeasible problem.}
	\label{fig:featuresLR_compas}
\end{figure}

Next, we display for all the instances in the {\tt COMPAS} dataset that were classified in the negative class by the logistic regression model, i.e.\ that comply with \eqref{eq:x0neg}, the key features that globally would need to be modified in order to change its class. We solve Problem \eqref{eq:CELRhard} for $F_{\text{max}}=1,2,3,4,5$. For $F_{\text{max}}=1$, the optimization problem is infeasible. For the rest of the values of $F_max$, the features that are changed are displayed in Figure \ref{fig:featuresLR_compas}.

We reach the same conclusion as before: the only critical features to change the predicted class are AgeGroup and PriorsCount. 

\begin{figure}[h]
	\subfloat[$\lambda_{glob}=0.1$\label{fig:outlier01_compas}]
	{\includegraphics[width=.35\linewidth]{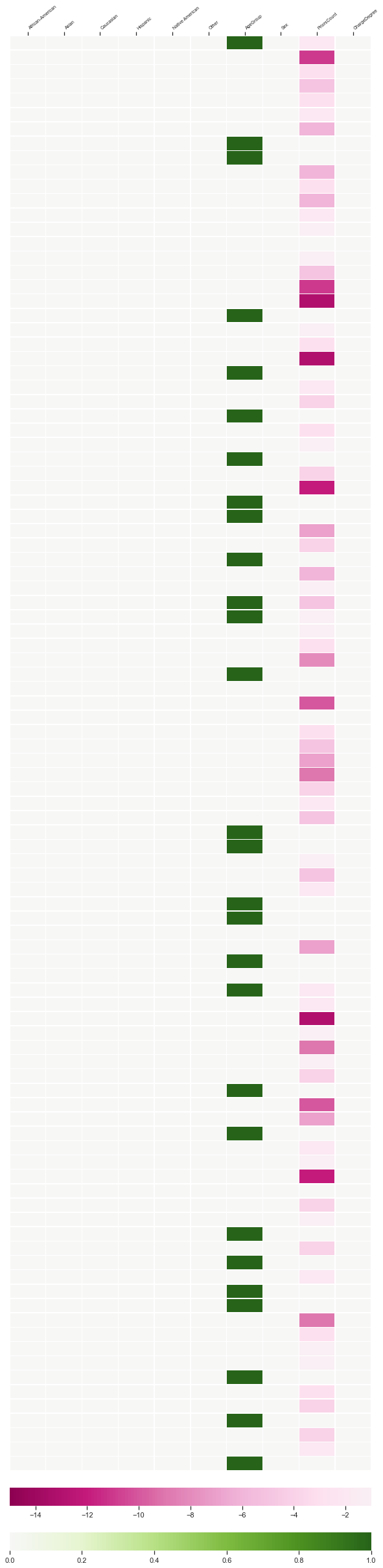}}\hfill
	\subfloat[$\lambda_{glob}=10$\label{figoutlier10_compas}]
	{\includegraphics[width=.35\linewidth]{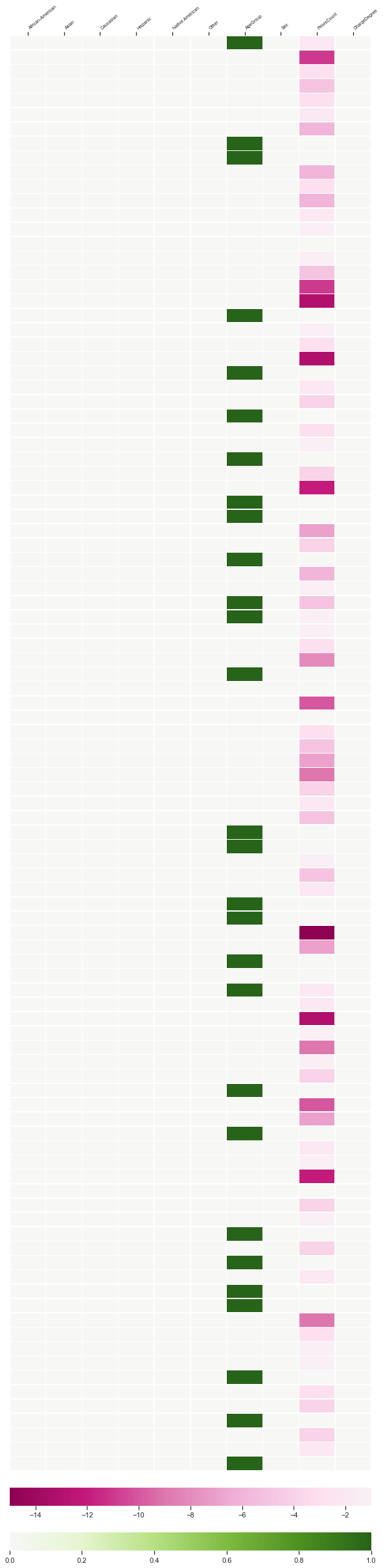}}
	\caption{Visualization of 100 of the counterfactual explanations for all the instances in the {\tt COMPAS} dataset that comply with \eqref{eq:x0neg}. The classifier considered is a logistic regression model. The explanations have been calculated solving model \eqref{eq:CELR} with $\lambda_{ind}=0$, $\lambda_{glob}=0.1$ (left), $10$ (right), and $I^{*}=\lceil 0.95I \rceil$. The feature perturbations are displayed. }
	\label{fig:outliers_compas}
\end{figure}

Finally, for this dataset, we look at $\mathcal{I}^*$, which can detect outliers from a group. The counterfactuals can be seen in Figure \ref{fig:outliers_compas}.

For both values of lambda, the two features that are perturbed are the same. We obtain two optimal sets of outliers with the same total euclidean distance between the original samples and the counterfactuals. 

\subsection*{Credit}

The {\tt{Credit}} dataset \citep{yeh2009comparisons}  was used to predict the probability of default payments in Taiwan. The data was preprocessed as in \cite{karimi2020model, parmentier2021optimal}. The description of the dataset is detailed in Table \ref{tab:features_credit}. 

\begin{table}[h]
	\centering
	\resizebox{1\columnwidth}{!}{%
		\begin{tabular}{c|c|c}
			Variable & Definition & Type \\ \hline
			isMale   &  sex of individual: 0 (female), 1(male) & binary\\
			isMarried   &  marital status: 0 (not married), 1 (married) &binary\\
			AgeGroup & age category: 1 ($<$25), 2 (25-39), 3 (40-59), 4 ($\geq$60) & categorical (ordinal)\\
			EducationLevel & education level: 1 (other), 2 (highschool), 3 (university), 4 (graduate) & categorical (ordinal)\\
			MaxBillAmountOverLast6Months & maximum bill amount over the last 6 months & numerical\\
			MaxPaymentAmountOverLast6Months & maximum payment amount over the last 6 months & numerical\\
			MonthsWithZeroBalanceOverLast6Months & months with zero balance over the last 6 months & numerical\\
			MonthsWithLowSpendingOverLast6Months & months with low spending (bill$<$0.2) over last 6 months & numerical\\
			MonthsWithHighSpendingOverLast6Months & months with high spending (bill$>$0.8) over last 6 months & numerical\\
			MostRecentBillAmount & most recent bill amount & numerical\\
			MostRecentPaymentAmount & most recent payment amount & numerical\\
			TotalOverdueCounts & total overdue counts & numerical\\
			TotalMonthsOverdue & total months overdue & numerical\\
			HasHistoryOfOverduePayments & binary variable indicating whethere the individual has a history of overdue payment: yes(1), no (0) & binary \\
			NoDefaultNextMonth & binary variable indicating whether the individual will default next month (0) or not (1) & target\\
	\end{tabular}}%
	\caption{Description of the features of the {\tt Credit} dataset}
	\label{tab:features_credit}
\end{table}

First, counterfactual explanations for 10 individuals that were classified originally in the negative class by both classifiers, are calculated. The original values of these 10 instances can be seen in Table \ref{tab:x0values_credit}.

\begin{sidewaystable}[t]
	\centering
	\resizebox{\textwidth}{!}{%
		\begin{tabular}{cccccccccccccc}
			\toprule
			isMale & isMarried & AgeGroup & EducationLevel &  MaxBillAmountOverLast6Months &  MaxPaymentAmountOverLast6Months & MonthsWithZeroBalanceOverLast6Months & MonthsWithLowSpendingOverLast6Months & MonthsWithHighSpendingOverLast6Months &  MostRecentBillAmount &  MostRecentPaymentAmount & TotalOverdueCounts & TotalMonthsOverdue & HasHistoryOfOverduePayments \\
			\midrule
			1 &         1 &   3 &              4 &                        0.1704 &                           0.0107 &                                    0 &                                    0 &                                     6 &                0.2659 &                   0.0357 &                  1 &                 22 &                           1 \\
			1 &         0 &   2 &              3 &                        0.0632 &                           0.0119 &                                    0 &                                    0 &                                     5 &                0.0754 &                   0.0396 &                  1 &                 32 &                           1 \\
			1 &         0 &   2 &              2 &                        0.1431 &                           0.0169 &                                    0 &                                    0 &                                     6 &                0.2217 &                   0.0201 &                  1 &                 12 &                           1 \\
			0 &         1 &   2 &              4 &                        0.2153 &                           0.0095 &                                    0 &                                    0 &                                     6 &                0.3341 &                   0.0318 &                  1 &                 12 &                           1 \\
			1 &         0 &   2 &              4 &                        0.1724 &                           0.0072 &                                    0 &                                    0 &                                     6 &                0.2832 &                   0.0240 &                  2 &                  8 &                           1 \\
			1 &         1 &   3 &              4 &                        0.0811 &                           0.0087 &                                    0 &                                    0 &                                     5 &                0.1175 &                   0.0292 &                  2 &                  8 &                           1 \\
			1 &         0 &   2 &              4 &                        0.1079 &                           0.0089 &                                    0 &                                    0 &                                     5 &                0.1623 &                   0.0298 &                  1 &                 12 &                           1 \\
			1 &         1 &   3 &              2 &                        0.1559 &                           0.0086 &                                    0 &                                    0 &                                     6 &                0.2689 &                   0.0220 &                  1 &                 12 &                           1 \\
			1 &         0 &   2 &              4 &                        0.1071 &                           0.0086 &                                    0 &                                    0 &                                     6 &                0.1599 &                   0.0285 &                  1 &                  8 &                           1 \\
			1 &         0 &   2 &              3 &                        0.1785 &                           0.0078 &                                    0 &                                    0 &                                     6 &                0.2958 &                   0.0259 &                  1 &                 12 &                           1 \\
			\bottomrule
		\end{tabular}%
	}
	\caption{Feature values of $\bm{\underline {x}}^0=(\bm{x}^0_1,\dots,\bm{x}^0_{10})$ used for the models in Figures \ref{fig:CEsep_credit}--\ref{fig:CEgroup_credit}.}
	\label{tab:x0values_credit}
\end{sidewaystable}

We solve models \eqref{eq:CEsep}, \eqref{eq:CELR} and \eqref{eq:CEATM}, considering different combinations of the values of $\lambda_{ind}$ and $\lambda_{glob}$.

\begin{figure}[H]
	\subfloat[Logistic regression\label{fig:indLR_credit}]
	{\includegraphics[width=.48\linewidth]{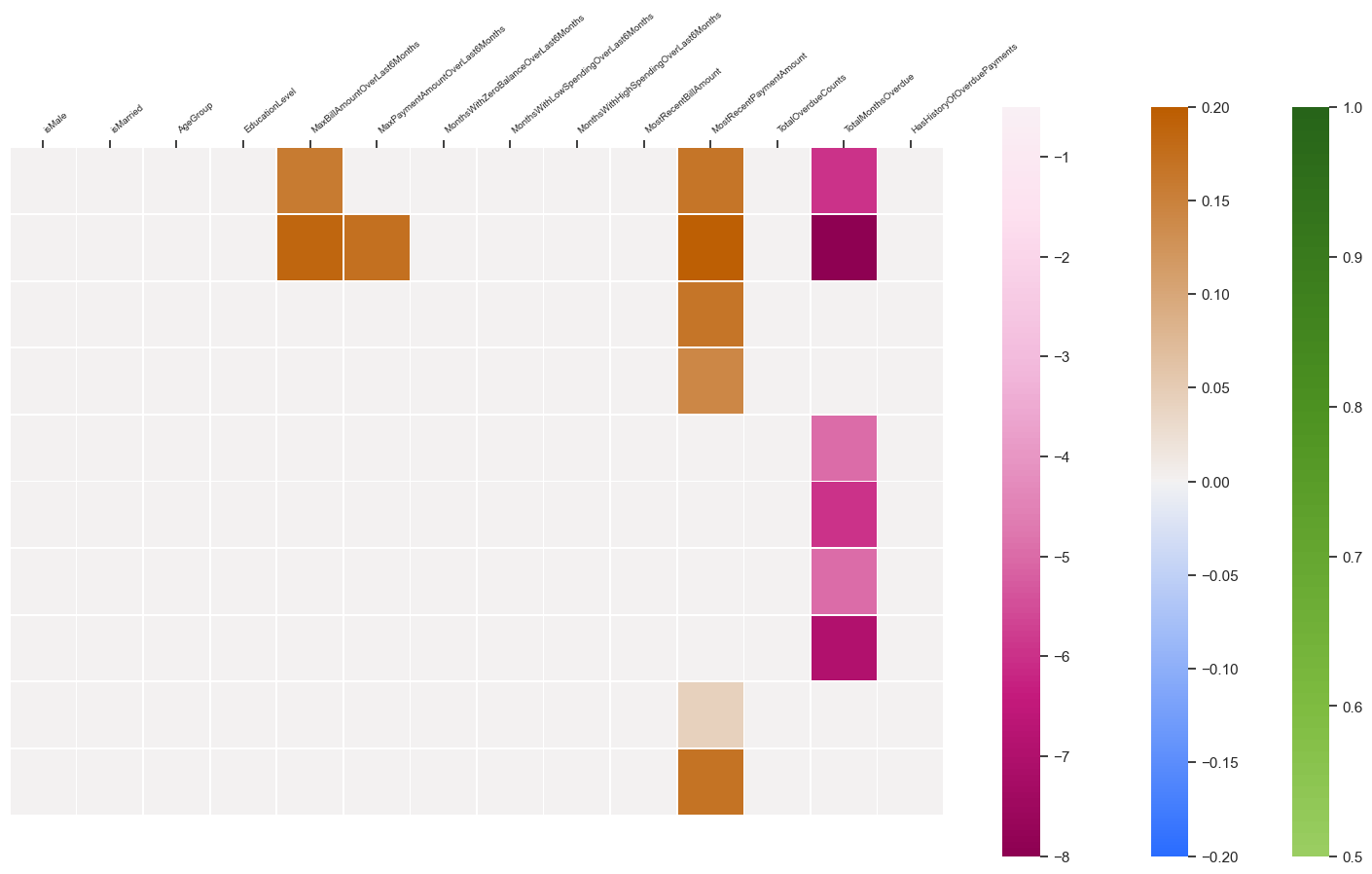}}\hfill
	\subfloat[Random forest\label{fig:indRF_credit}]
	{\includegraphics[width=.48\linewidth]{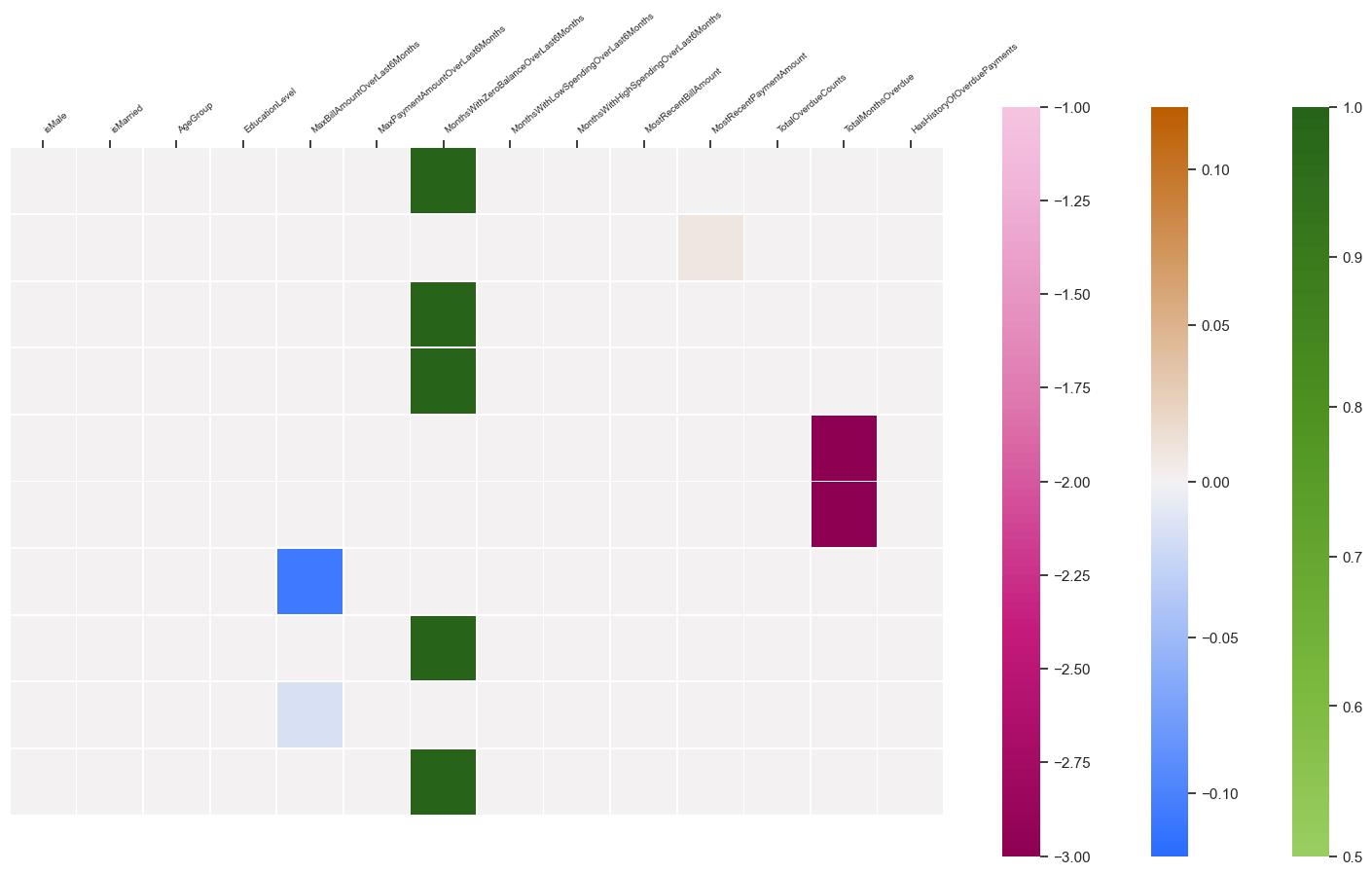}}
	\caption{Counterfactual explanations for instances $\bm{x}_i^0$ in Table \ref{tab:x0values_credit}
		for two classifiers, the logistic regression and the random forest. The explanations have been calculated solving model \eqref{eq:CEsep} with $C_i(\bm{x}_i^0,\bm{x}_i)= \|\bm{x}^0_i-\bm{x}_i\|_2^2 +  \lambda_{ind}  \|\bm{x}_i^0-\bm{x}_i\|_0$ for $\lambda_{ind}=0.02$. The feature perturbations are displayed.}
	\label{fig:CEsep_credit}
\end{figure}

\begin{figure}[H]
	\subfloat[Logistic regression\label{fig:globLR_credit}]
	{\includegraphics[width=.45\linewidth]{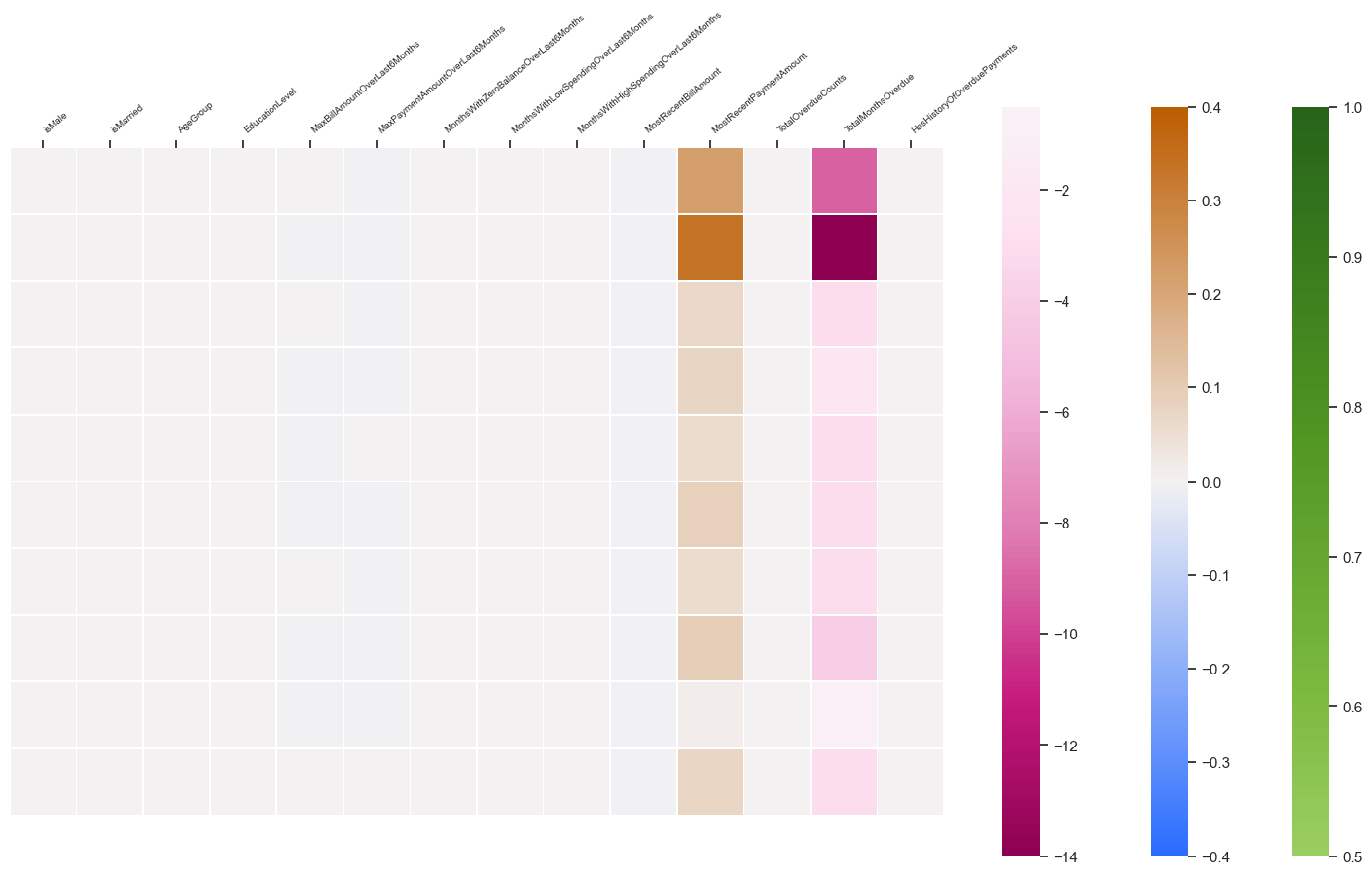}}\hfill
	\subfloat[Random forest\label{fig:globRF_credit}]
	{\includegraphics[width=.45\linewidth]{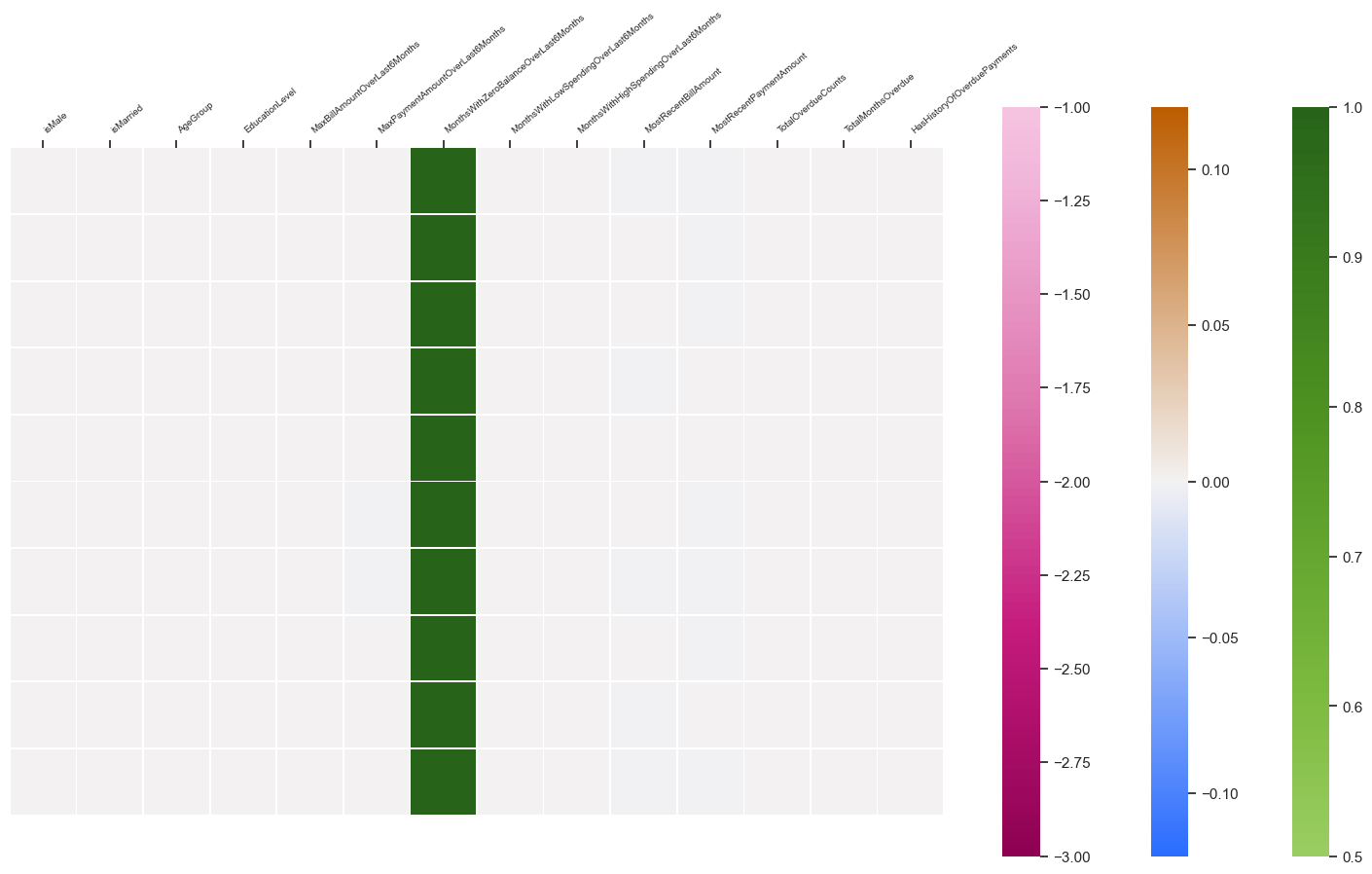}}
	\caption{Counterfactual explanations for instances $\bm{x}_i^0$ in Table \ref{tab:x0values_credit}
		for two classifiers, the logistic regression and the random forest. The explanations have been calculated solving models \eqref{eq:CELR} and \eqref{eq:CEATM} with cost function \eqref{eq:generalC} with $\lambda_{ind}=0$, $\lambda_{glob}=0.2$. The feature perturbations are displayed.   }
	\label{fig:CEgroup_credit}
\end{figure}

For this dataset, to not be labeled as default, the number of total months overdue and the most recent payment amount has to decrease, whereas the number of months with zero balance has to increase. 

Comparing the changes in the separable case (Figure \ref{fig:CEsep_credit}) with the non-separable case (Figure \ref{fig:CEgroup_credit}), we can see a big difference on the number of features needed to be perturbed. We see that one must change 4 features globally in the separable case, both in the logistic regression and random forest case, whereas it is reduced to 2 and 1 feature in the non-separable case, respectively. 

Looking at the results, it can be seen that the critical features are different for each classifier for these 10 instances. For the logistic regression, the features that have the greatest impact are MostRecentPayment and TotalMonthsOverdue, whereas for the random forest is the feature about the number of months with zero balance over the last 6 months.

\begin{figure}[H]
	\centering	
	\includegraphics[scale=.5]{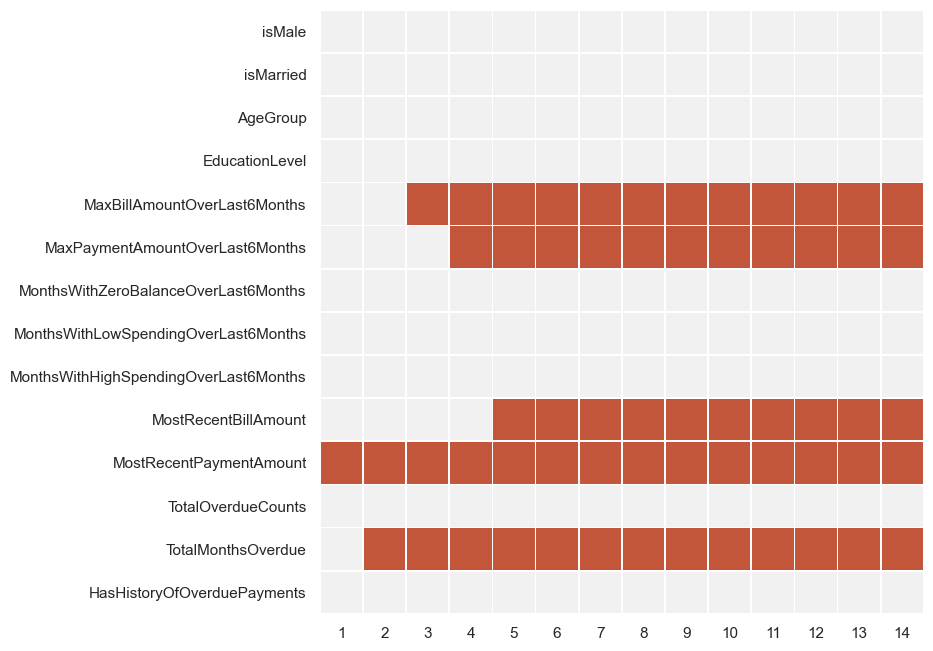}
	\caption{Features that need to be perturbed (in red) for all the instances in the {\tt Credit} dataset that comply with \eqref{eq:x0neg} when solving Problem \eqref{eq:CELRhard} for all the values of $F_{\text{max}}$ to describe the Pareto front. The classifier considered is a logistic regression model.}
	\label{fig:featuresLR_credit}
\end{figure}

Next, we display for all the instances in the {\tt Credit} dataset that comply with \eqref{eq:x0neg}, the key features that globally would need to to be modified in order to change its class. We solve Problem \eqref{eq:CELRhard} for the different values of $F_{\text{max}}$ from 1 to 14. The features changed are displayed in Figure \ref{fig:featuresLR_credit}.

We can see that the maximum number of features needed to change the prediction are 5. 

\begin{figure}[h]
	\subfloat[$\lambda_{glob}=0.1$\label{fig:outlier01_credit}]
	{\includegraphics[width=.35\linewidth]{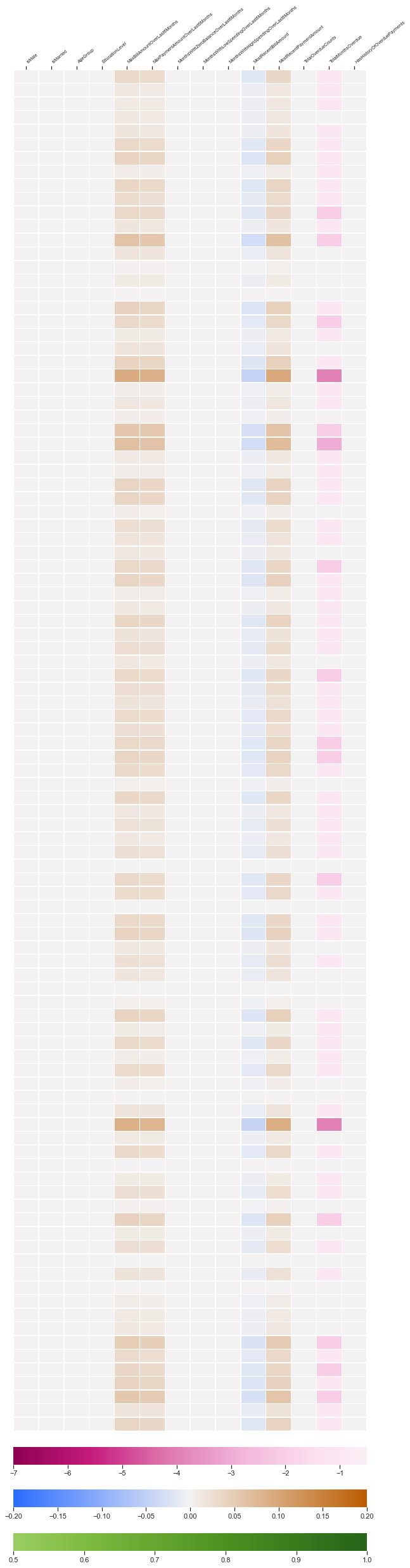}}\hfill
	\subfloat[$\lambda_{glob}=10$\label{figoutlier10_credit}]
	{\includegraphics[width=.35\linewidth]{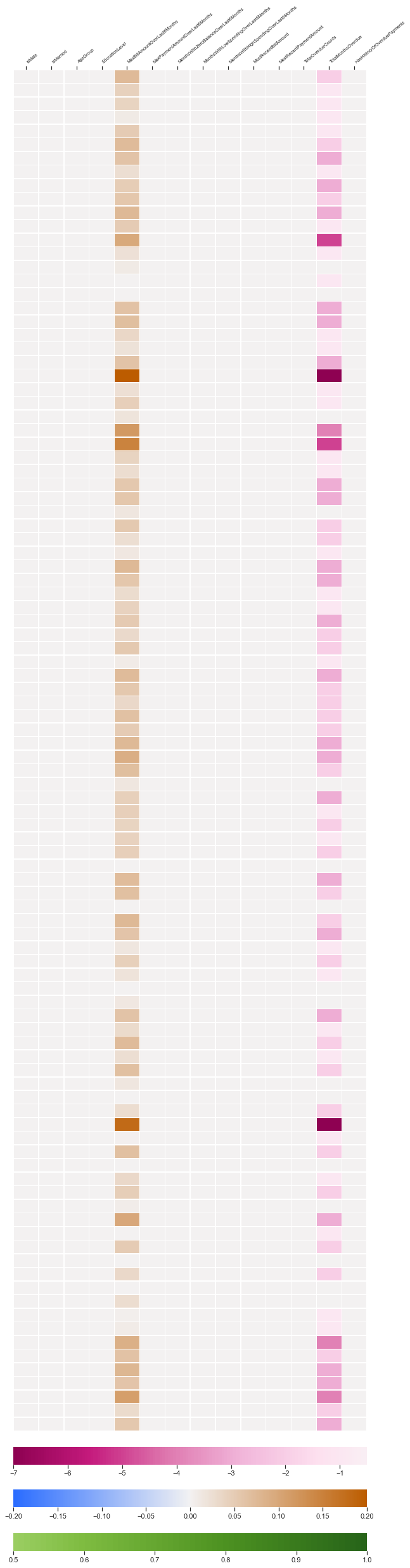}}
	\caption{Visualization of 100 of the counterfactual explanations for all the instances in the {\tt Credit} dataset that comply with \eqref{eq:x0neg}. The classifier considered is a logistic regression model. The explanations have been calculated solving model \eqref{eq:CELR} with $\lambda_{ind}=0$, $\lambda_{glob}=0.1$ (left), $10$ (right), and $I^{*}=\lceil 0.95I \rceil$. The feature perturbations are displayed. }
	\label{fig:outliers_credit}
\end{figure}

Finally, for this dataset, we look at $\mathcal{I}^*$, which can detect outliers from a group. The counterfactuals can be seen in Figure \ref{fig:outliers_credit}.

From the 144 outliers, 113 are in common for both cost functions.

\subsection*{Student's performance}

The {\tt{Students Performance}} dataset includes student grades, demographic, social and school related features and aims at predicting whereas a student will pass a course or not. The description of the dataset is detailed in Table \ref{tab:features_students}. 

\begin{table}[h]
	\centering
	\resizebox{1\columnwidth}{!}{%
		\begin{tabular}{c|c|c}
			Variable & Definition & Type \\ \hline
			school   & student's school: 0 (Gabriel Pereira), 1 (Mousinho da Silveira) & binary\\
			sex   &  student's sex: 0 (male), 1 (female) &binary\\
			age & student's age & numerical\\
			address  & student's home address type: 0 (rural), 1(urbal) & binary\\
			famsize & family size: 0 (less or equal to 3), 1 (greater than 3) & binary\\
			Pstatus & parent's cohabitation status: 0 (living together), 1 (apart) & binary\\
			Medu  & mother's education: 0 (none), 1 (primary education), 2 (5th to 9th grade), 3 (secondary education), 4 (higher education) & categorical (ordinal)\\
			Fedu  & father's education:  0 (none), 1 (primary education), 2 (5th to 9th grade), 3 (secondary education), 4 (higher education)  & categorical (ordinal)\\
			Mjob  &  mother's job: 'teacher', 'health' care related, civil 'services' (e.g. administrative or police), 'at\_home' or 'other' & categorical (nominal)\\
			Fjob & father's job: 'teacher', 'health' care related, civil 'services' (e.g. administrative or police), 'at\_home' or 'other' & categorical (nominal)\\
			reason & reason to choose this school: close to 'home', school 'reputation', 'course' preference or 'other' & categorical (nominal)\\
			guardian  & student's guardian: 'mother', 'father' or 'other' & categorical (nominal)\\
			traveltime  & home to school travel time: 1 ($<$15 min) , 2 (15 to 30 min), 3 (30 min to 1 hour), 4 ($>$1 hour) & categorical (ordinal)\\
			studytime &  weekly study time: 1 ($<$2 hours), 2 (2 to 5 hours), 3 (5 to 10 hours), or 4 ($>$10 hours) & categorical (ordinal) \\
			failures  & number of past class failures & numerical\\
			schoolsup & extra educational support: 0 (no), 1 (yes) & binary\\
			famsup & family educational support: 0 (no), 1 (yes) & binary\\
			paid & extra paid classes within the course subject: 0 (no), 1 (yes) & binary\\
			activities & extra-curricular activities: 0 (no), 1 (yes) & binary \\
			nursery & attended nursery school: 0 (no), 1 (yes) & binary \\
			higher & wants to take higher education : 0 (no), 1 (yes) & binary\\
			internet &  Internet access at home: 0 (no), 1 (yes) & binary\\
			romantic &  with a romantic relationship: 0 (no), 1 (yes) & binary\\
			famrel & quality of family relationships: from 1 (very bad) to 5(excellent) & categorical (ordinal)\\
			freetime &  free time after school: from 1 (very low) to 5 (very high) & categorical (ordinal) \\
			goout & going out with friends: from 1 (very low) to 5 (very high) & categorical (ordinal) \\
			Dalc & workday alcohol consumption: from 1 (very low) to 5 (very high) & categorical (ordinal) \\
			Walc &weekend alcohol consumption: from 1 (very low) to 5 (very high) & categorical (ordinal) \\
			health &current health status: : from 1 (very low) to 5 (very high) & categorical (ordinal) \\
			absences& number of school absences & numerical \\
			Class & 1 if the student passes the math course, -1 otherwise & target	
	\end{tabular}}%
	\caption{Description of the features of the {\tt Students performance} dataset}
	\label{tab:features_students}
\end{table}

First, counterfactual explanations for 10 individuals that were classified originally in the negative class by both classifiers, are calculated. The original values of these 10 instances can be seen in Table \ref{tab:x0values_students}.

\begin{sidewaystable}[t]
	\centering
	\resizebox{\textwidth}{!}{%
		\begin{tabular}{cccccccccccccccccccccccccccccc}
			\toprule
			Mjob &    Fjob &     reason & guardian & school & sex & age & address & famsize & Pstatus & Medu & Fedu & traveltime & studytime & failures & schoolsup & famsup & paid & activities & nursery & higher & internet & romantic & famrel & freetime & goout & Dalc & Walc & health & absences \\
			\midrule
			services &   other & reputation &   father &      0 &   1 &  15 &       1 &       1 &       0 &    2 &    1 &          3 &         3 &        0 &         1 &      1 &    1 &          1 &       1 &      1 &        1 &        1 &      5 &        2 &     2 &    1 &    1 &      4 &        4 \\
			at\_home &   other & reputation &   mother &      0 &   1 &  15 &       0 &       1 &       0 &    2 &    2 &          1 &         1 &        0 &         1 &      1 &    1 &          1 &       1 &      1 &        1 &        1 &      4 &        3 &     1 &    1 &    1 &      2 &        8 \\
			other & at\_home &     course &   father &      0 &   1 &  16 &       1 &       0 &       0 &    2 &    2 &          2 &         2 &        1 &         1 &      1 &    1 &          1 &       1 &      1 &        1 &        1 &      4 &        3 &     3 &    2 &    2 &      5 &       14 \\
			other &   other &       home &   mother &      0 &   1 &  16 &       1 &       1 &       0 &    2 &    2 &          1 &         2 &        0 &         1 &      1 &    1 &          1 &       1 &      1 &        1 &        1 &      5 &        4 &     4 &    1 &    1 &      5 &        0 \\
			teacher &   other &     course &   mother &      0 &   0 &  17 &       1 &       0 &       0 &    4 &    3 &          2 &         2 &        0 &         1 &      1 &    1 &          1 &       1 &      1 &        1 &        1 &      4 &        4 &     4 &    4 &    4 &      4 &        4 \\
			other &   other & reputation &   mother &      0 &   1 &  16 &       1 &       1 &       0 &    2 &    3 &          1 &         2 &        0 &         1 &      1 &    1 &          1 &       1 &      1 &        1 &        1 &      4 &        4 &     3 &    1 &    3 &      4 &        6 \\
			other &   other & reputation &   mother &      0 &   1 &  18 &       0 &       1 &       0 &    3 &    1 &          1 &         2 &        1 &         1 &      1 &    1 &          1 &       1 &      1 &        1 &        1 &      5 &        3 &     3 &    1 &    1 &      4 &       16 \\
			other &   other &     course &   father &      0 &   0 &  17 &       1 &       1 &       0 &    2 &    3 &          2 &         1 &        0 &         1 &      1 &    1 &          1 &       1 &      1 &        1 &        1 &      5 &        2 &     2 &    1 &    1 &      2 &        4 \\
			health &   other &       home &   mother &      0 &   1 &  18 &       1 &       0 &       1 &    4 &    4 &          1 &         2 &        0 &         1 &      1 &    1 &          1 &       1 &      1 &        1 &        1 &      4 &        2 &     4 &    1 &    1 &      4 &       14 \\
			other &   other &     course &   mother &      1 &   0 &  18 &       0 &       1 &       0 &    3 &    2 &          2 &         1 &        1 &         1 &      1 &    1 &          1 &       1 &      1 &        1 &        1 &      2 &        5 &     5 &    5 &    5 &      5 &       10 \\
			\bottomrule
		\end{tabular}%
	}
	\caption{Feature values of $\bm{\underline {x}}^0=(\bm{x}^0_1,\dots,\bm{x}^0_{10})$ used for the models in Figures \ref{fig:CEsep_students}--\ref{fig:CEgroup_students}.}
	\label{tab:x0values_students}
\end{sidewaystable}

We solve models \eqref{eq:CEsep}, \eqref{eq:CELR} and \eqref{eq:CEATM}, considering different combinations of the values of $\lambda_{ind}$ and $\lambda_{glob}$.

\begin{figure}[H]
	\subfloat[Logistic regression\label{fig:indLR_students}]
	{\includegraphics[width=\linewidth]{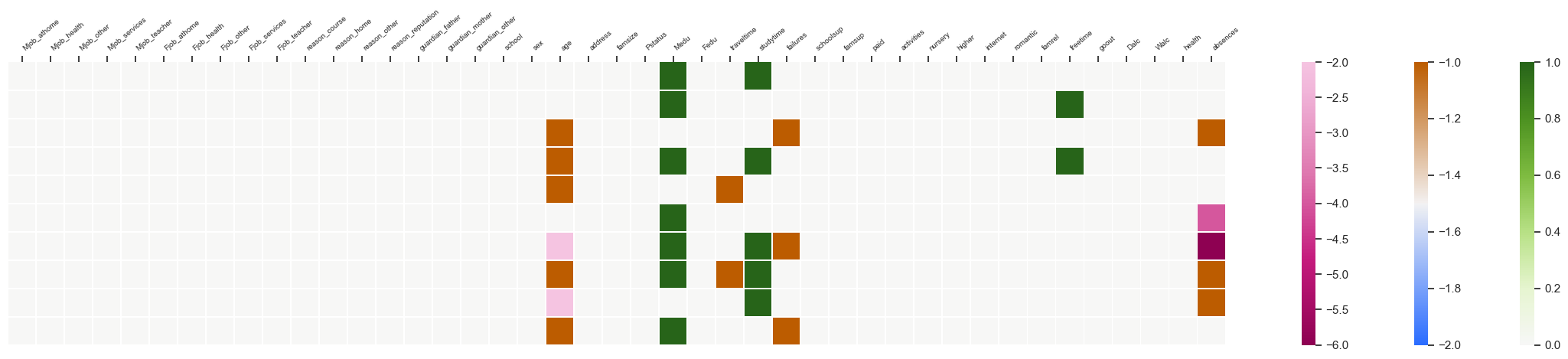}}\hfill
	\subfloat[Random forest\label{fig:indRF_students}]
	{\includegraphics[width=\linewidth]{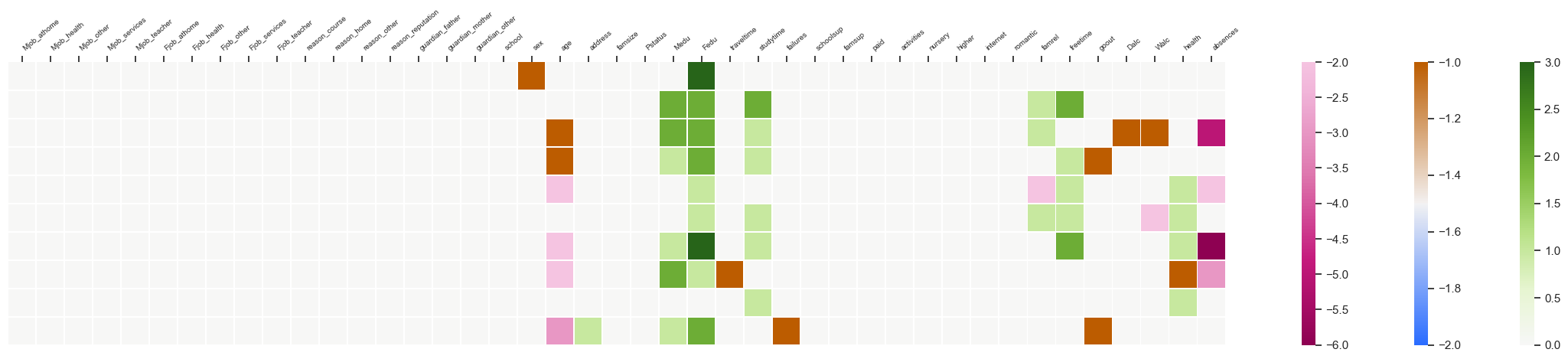}}
	\caption{Counterfactual explanations for instances $\bm{x}_i^0$ in Table \ref{tab:x0values_students}
		for two classifiers, the logistic regression and the random forest. The explanations have been calculated solving model \eqref{eq:CEsep} with $C_i(\bm{x}_i^0,\bm{x}_i)= \|\bm{x}^0_i-\bm{x}_i\|_2^2 +  \lambda_{ind}  \|\bm{x}_i^0-\bm{x}_i\|_0$ for $\lambda_{ind}=0.02$. The feature perturbations are displayed.}
	\label{fig:CEsep_students}
\end{figure}

\begin{figure}[H]
	\subfloat[Logistic regression\label{fig:globLR_students}]
	{\includegraphics[width=\linewidth]{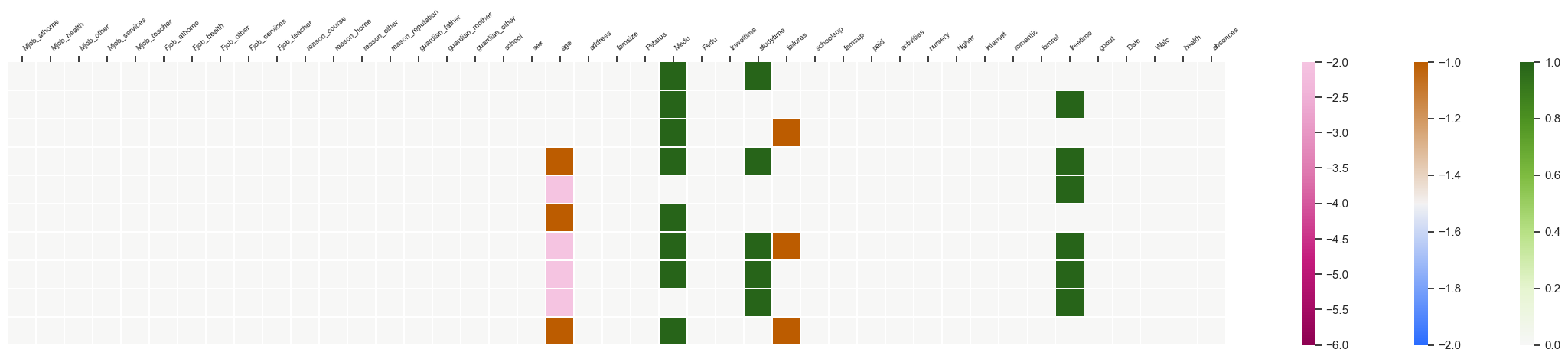}}\hfill
	\subfloat[Random forest\label{fig:globRF_students}]
	{\includegraphics[width=\linewidth]{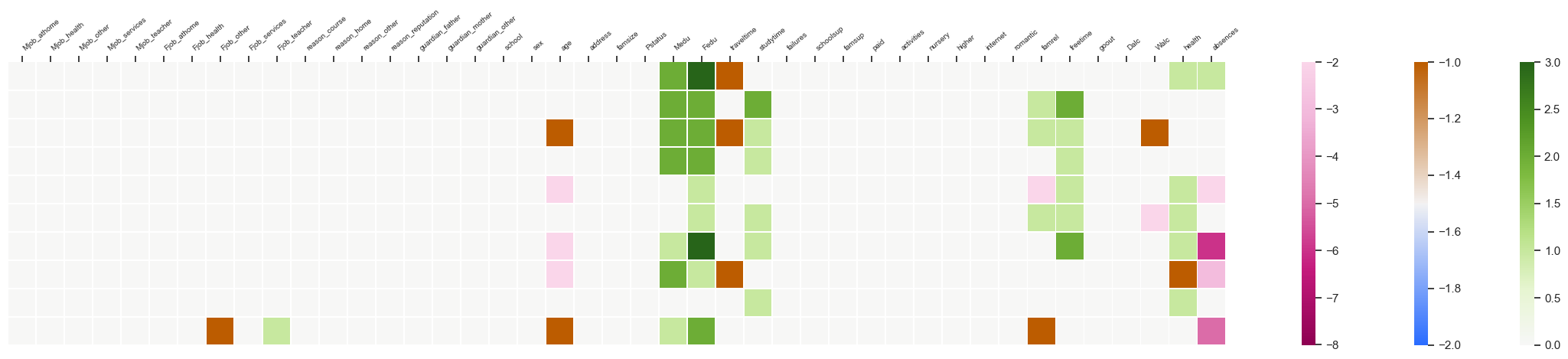}}
	\caption{Counterfactual explanations for instances $\bm{x}_i^0$ in Table \ref{tab:x0values_students}
		for two classifiers, the logistic regression and the random forest. The explanations have been calculated solving models \eqref{eq:CELR} and \eqref{eq:CEATM} with cost function \eqref{eq:generalC} with $\lambda_{ind}=0$, $\lambda_{glob}=0.2$. The feature perturbations are displayed.   }
	\label{fig:CEgroup_students}
\end{figure}

For this dataset, to be classified in the positive class, i.e., passing the course, the features that have to be increased are mostly the mother's and father's education, the study time, the free time, the quality of family relations and the health status; whereas the features that have to be mostly decreased are the age, the number of failures, the number of absences, travel time and how much they go out with friends. 

Comparing the changes in the separable case (Figure \ref{fig:CEsep_students}) with the non-separable case (Figure \ref{fig:CEgroup_students}), one must perturb 7 and 15 features globally in the separable case in the logistic regression and random forest, respectively. This is reduced to 5 and 11 features in the non-separable case.

Focusing in the non-separable case ,we clearly see in the figures that the features that have the greatest impact are the age of the students, the mother's education, the study time, the number of failures and amount of free time that the students have. In the case of the random forest there seems to be more features that impact on the decision made by the classifiers. Besides the one already mentioned, the father's education, the health status, the travel time and the number of failures are also perturbed for most records.

Next, we display for all the instances in the {\tt Students performance} dataset that comply with \eqref{eq:x0neg}, the key features that globally would need to be perturbed in order to change its class. We solve Problem \eqref{eq:CELRhard} for different values of the parameter $F_{\text{max}},$ ranging from 1 to 30. For values 1 to 5, the problem was infeasible. The features changed are displayed in Figure \ref{fig:featuresLR_credit}.

\begin{figure}[H]
	\centering	
	\includegraphics[scale=.5]{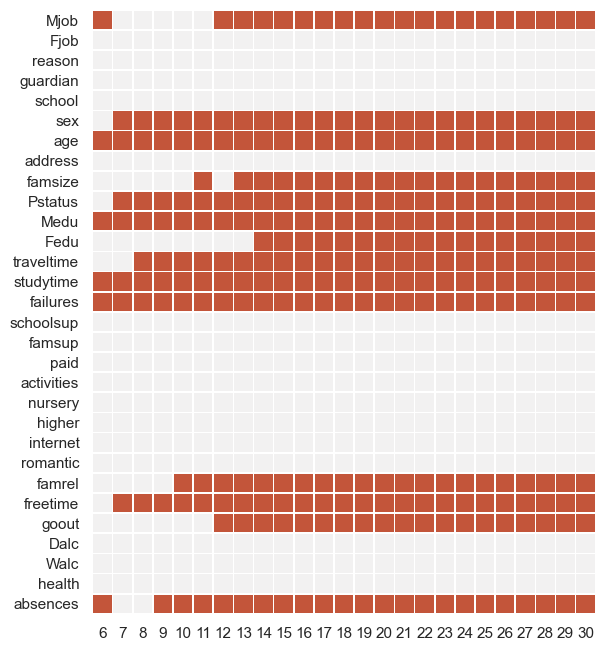}
	\caption{Features that need to be perturbed (in red) for all the instances in the {\tt Students performance} dataset that comply with \eqref{eq:x0neg} when solving Problem \eqref{eq:CELRhard} for all the values of $F_{\text{max}}$ to describe the Pareto front. The classifier considered is a logistic regression model. $F_{\text{max}}=1,2,3,4,5$ yields an infeasible problem.}
	\label{fig:featuresLR_students}
\end{figure}

We can see that the maximum number of features needed to change the prediction are 14. 

\begin{figure}[h]
	\subfloat[$\lambda_{glob}=0.1$\label{fig:outlier01_students}]
	{\includegraphics[width=.35\linewidth]{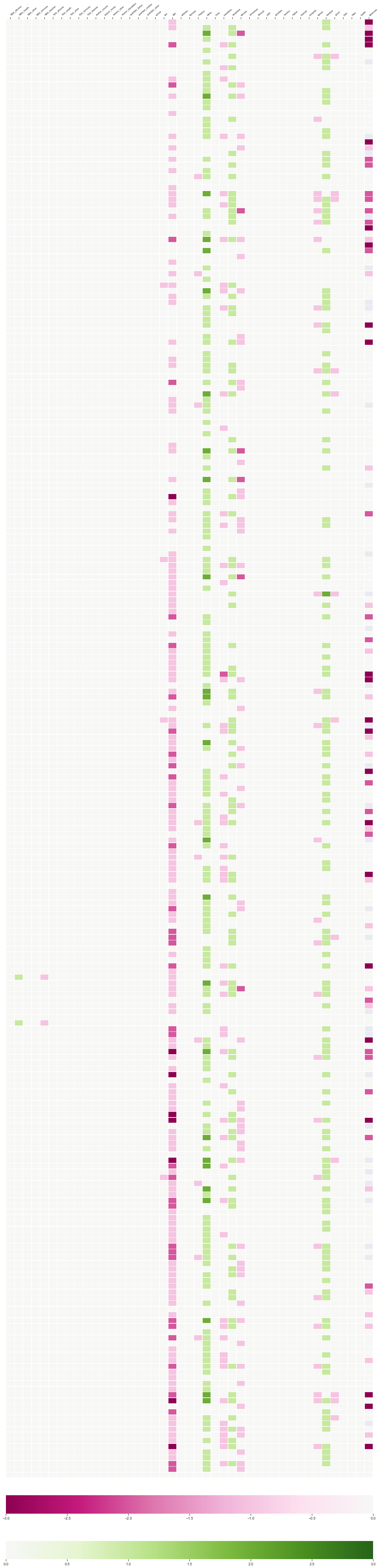}}\hfill
	\subfloat[$\lambda_{glob}=10$\label{figoutlier10_students}]
	{\includegraphics[width=.35\linewidth]{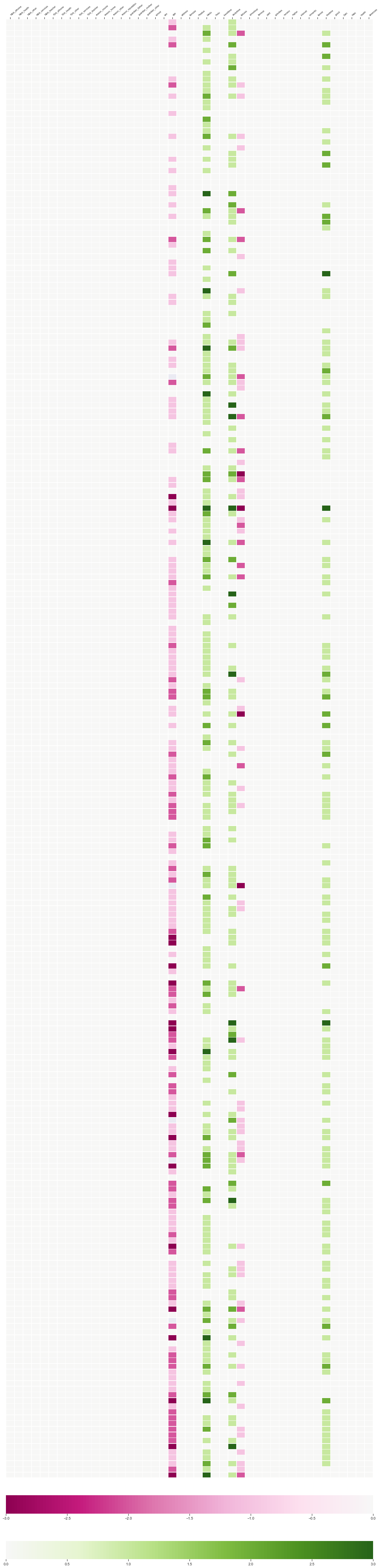}}
	\caption{Counterfactual explanations for all the instances in the {\tt Students performance} dataset that comply with \eqref{eq:x0neg}. The classifier considered is a logistic regression model. The explanations have been calculated solving model \eqref{eq:CELR} with $\lambda_{ind}=0$, $\lambda_{glob}=0.1$ (left), $10$ (right), and $I^{*}=\lceil 0.95I \rceil$. The feature perturbations are displayed. }
	\label{fig:outliers_students}
\end{figure}

Finally, for this dataset, we look at $\mathcal{I}^*$, which can detect outliers from a group. The counterfactuals can be seen in Figure \ref{fig:outliers_students}.

For $\lambda_{glob}=0.1$ the indices of the outliers are 28, 57, 62, 69, 79, 85, 91, 121, 151, 174, 198, 225, 254, whereas for $\lambda_{glob}=10$ the indices of the outliers are 27, 28, 31, 46, 50, 122, 124, 140, 146, 167, 174, 202, 216. There are only two in common. 

\end{document}